\documentclass[twoside]{article}

\usepackage[accepted]{aistats2025}
\usepackage[utf8]{inputenc} 

\usepackage{hyperref}       
\usepackage{url}            
\usepackage{booktabs}       
\usepackage{nicefrac}       
\usepackage{microtype}      
\usepackage{xcolor}         

\usepackage[T1]{fontenc}    
\usepackage{amsfonts}       

\usepackage{mathtools} 
\usepackage{booktabs} 
\usepackage{overpic}

\usepackage{lipsum}

\usepackage{ctable} 
\usepackage{makecell}
\usepackage{amsmath,amssymb}
\usepackage{subcaption}
\usepackage{float}
\usepackage{multirow}
\usepackage{array}
\newcolumntype{?}{!{\vrule width 1.5pt}} 

\begin{document}

%

%

\twocolumn[

\aistatstitle{A Multi-Task Learning Approach to Linear Multivariate Forecasting}

\aistatsauthor{ Liran Nochumsohn \And Hedi Zisling \And  Omri Azencot }

\aistatsaddress{\And  Ben-Gurion University of the Negev \And} ]

\begin{abstract}
Accurate forecasting of multivariate time series data is important in many engineering and scientific applications. Recent state-of-the-art works ignore the inter-relations between variates, using their model on each variate independently. This raises several research questions related to proper modeling of multivariate data. In this work, we propose to view multivariate forecasting as a \emph{multi-task learning} problem, facilitating the analysis of forecasting by considering the angle between task gradients and their balance. To do so, we analyze linear models to characterize the behavior of tasks. Our analysis suggests that tasks can be defined by grouping similar variates together, which we achieve via a simple clustering that depends on correlation-based similarities. Moreover, to balance tasks, we scale gradients with respect to their prediction error. Then, each task is solved with a linear model within our \emph{MTLinear} framework. We evaluate our approach on challenging benchmarks in comparison to strong baselines, and we show it obtains on-par or better results on multivariate forecasting problems. Code is available at \href{https://github.com/azencot-group/MTLinear}{https://github.com/azencot-group/MTLinear}. 

\end{abstract}

\section{INTRODUCTION}
\label{sec:intro}

Time series forecasting (TSF) with deep learning leverages neural networks to model and predict sequential data over time, enabling accurate predictions and insights into future trends. Its importance lies in its ability to handle complex temporal dependencies and patterns, making it invaluable for tasks such as financial forecasting, resource planning, and demand prediction in various industries. While most existing TSF approaches are based on N-BEATS and the Transformer~\cite{oreshkin2020nbeats, vaswani2017attention}, a recent work finds simple linear layers to be highly effective \cite{zeng2023Transformers, li2023revisiting}. However, linear models are naturally limited, and thus, current efforts focus on developing nonlinear approaches where state-of-the-art (SOTA) techniques incorporate the linear module as the final decoder layer~\cite{nie2023time, zhou2023one}.

Generally, TSF frameworks are designed to accept univariate and multivariate temporal data. In the multivariate case, time series data has multiple dimensions per sequence sample, whereas univariate data is one dimensional. Remarkably, while TSF is assumed to benefit from the inter-relations underlying multivariate information \cite{granger1969investigating, schreiber2000measuring}, recent methods opt to handle each variate independently~\cite{zeng2023Transformers, nie2023time}. Nevertheless, dominant variate signals may disproportionally affect forecast performance, especially in limited linear models. Several research questions arise from the latter straightforward observation: how to model different variates? how to treat similar vs. dissimilar variates? how to balance the contribution of variates in the context of forecasting?

Towards addressing the above questions, we interpret time series forecasting with deep learning through the lens of multi-task learning (MTL)~\cite{zhang2021survey}. MTL trains a single model to perform multiple related tasks simultaneously, leveraging shared representations to improve performance across tasks, enable knowledge transfer, and facilitate efficient learning. Our view of TSF as MTL is based on the assumption that \emph{similar variates should be modeled similarly, and dissimilar variates encode different forecasting tasks.} When considered as separate tasks within MTL formalism, we can harness tools, observations, and the general advances in MTL to better solve time series forecasting problems.

Particularly, a recent work~\cite{yu2020gradient} formulates some of the challenges underlying multi-task learning with respect to the \emph{tragic triad}. The authors advocate that conflicting gradients, varying gradient magnitudes, and highly curved loss manifolds hinder MTL methodologies. Further, identifying which tasks can learn together is crucial to effective multi-task learning \cite{fifty2021efficiently}. For instance, assigning conflicting tasks to a separate set of network weights was suggested in~\cite{guangyuan2022recon}. In our study, we explore how to group variates together, and how to weigh different variate groups during training. To our knowledge, deep TSF from a multi-task learning perspective has received only limited attention. 

Given the success of linear models in TSF, we are motivated to study the use of one linear model per similar variates group, and to balance dissimilar variate groups based on their dominance. Specifically, we analyze the gradients of linear modules, and we observe the factors affecting the norm of the gradients and their direction. Based on our derivative analysis, we propose to group variates by their Pearson Correlation Coefficient, encoding linear relationships. Then, we construct a multi-head linear model (MTLinear), where each head learns from a separate variate group. Finally, we scale the gradients using the variate's characteristics.
Our contributions can be summarized as follows:
\begin{itemize}
    \item We suggest to interpret time series forecasting as multi-task learning, where similar variates are grouped together, and each group forms a separate task. Variate grouping and per group balancing is inspired by our gradients analysis of linear TSF.
    \item We propose a simple, efficient and effective multi-head linear network (\textbf{MTLinear}) for solving multivariate time series forecasting tasks.
    \item We extensively evaluate our method on challenging TSF benchmarks, and compare our results to state-of-the-art (SOTA) models, showing that MTLinear is a strong standalone technique and it can be considered as a building block module for TSF.
\end{itemize}

\section{BACKGROUND}
\label{sec:background}

\paragraph{Multi-task learning (MTL)}\hspace{-3mm} aims to optimize a set of weights $\theta$ that simultaneously minimize $k$ different tasks, each corresponding to a different loss function $L_i(\theta)$. A typical single-objective function that combines all tasks takes the following form:
\begin{equation} \label{eq:mtl}
\theta^* =\mathop{\arg \min}\limits_{\theta} L(\theta)  := \mathop{\arg \min}\limits_{\theta} \frac{1}{k} \sum_i^k L_i(\theta) \ ,
\end{equation}
where all tasks are averaged together to form a single loss $L(\theta)$ that can be directly minimized via gradient descent. In this setting, each loss $L_i(\theta)$ is associated with a gradient $g_i := \nabla L_i(\theta)$. The total gradient reads
\begin{equation} \label{eq:mtl_grad}
\nabla L(\theta) = \frac{1}{k} \sum_i^k g_i = \frac{1}{k} \sum_i^k \nabla L_i(\theta) \ .
\end{equation}
Following \cite{yu2020gradient}, we detail fundamental challenges inherent to multi-task learning, termed as the tragic triad. First, \emph{conflicting gradients} arise when the inner product between a pair of task gradients $g_i, g_j$ is negative, i.e., $g_i^T g_j < 0$. Geometrically, it means that their angle is greater than $\pi/2$, and therefore, optimizing $\theta$ is carried by non-aligned directions. Second, when a set of gradients have a large \emph{gradient magnitude difference}, larger gradients may shadow weaker gradients. This issue becomes especially problematic when gradients are also conflicting, leading to long and unstable training \cite{pascanu2013difficulty}. Finally, \emph{highly curved loss landscapes} negatively affect the optimization of neural networks~\cite{li2018visualizing, kaufman2023data, kaufman2024geometric}. 

\paragraph{Multivariate TSF}\hspace{-3mm} deals with predicting future values of multiple inter-related variates over the time domain. Particularly, we are given in the multivariate case $k$ inter-related components $x_1, x_2, \dots, x_k \in \mathbb{R}^l$ of length $l$, whereas the time series is univariate if $k=1$. In TSF we predict the corresponding future values, denoted by $y_1, y_2, \dots, y_k \in \mathbb{R}$. Formally, TSF solves
\begin{equation} \label{eq:tsf}
    \theta^* = \mathop{\arg \min}\limits_{\theta} F(\theta) := \mathop{\arg \min}\limits_{\theta} \frac{1}{k} \sum_i^k F_{i}(\theta) \ ,
\end{equation}
where the loss for forecasting tasks is the mean squared error (MSE), i.e., $F_i(\theta) := [M(x_i, \theta) - y_i]^2 \in \mathbb{R}$ with $M(\cdot, \cdot)$ being a parametric model such as a deep neural network or a linear module.

\paragraph{Linear forecasting}\hspace{-3mm} with Linear, NLinear, and DLinear in~\cite{zeng2023Transformers} includes a single linear layer. While extremely simple, these models have achieved remarkable results in comparison to many Transformer-based approaches. Yet, they fall short with respect to recent SOTA techniques~\cite{nie2023time, zhou2023one}. Notably, the latter works often include in their architectures a linear decoder, similar in structure and role to the linear layers proposed in~\cite{zeng2023Transformers}. Formally, a standard linear layer is defined as,
\begin{equation} \label{eq:linear_tsf}
\tilde{Y}^T = X^T \Theta + b \ ,
\end{equation}
where $X = [x_1, x_2, \dots, x_k] \in \mathbb{R}^{l \times k}$ is the input lookback, $\tilde{Y} = [y_1, y_2, \dots, y_k] \in \mathbb{R}^{h \times k}$ is the forecast horizon output, and $\Theta = [\theta_1, \theta_2, \dots, \theta_h] \in \mathbb{R}^{l \times h}$ and $b \in \mathbb{R}^h$ are the model weights and bias, respectively. We denote by $Y \in \mathbb{R}^{h \times k}$ the ground-truth horizon values.

\section{RELATED WORK}
\label{sec:related}

\paragraph{Multivariate time series forecasting.} Significant efforts have been dedicated to modeling and predicting sequential information using deep neural networks~\cite{goodfellow2016deep}. In particular, recurrent neural networks (RNNs) such as LSTM and GRU with their gating mechanisms obtained groundbreaking results on various vision and language tasks \cite{hochreiter1997long, cho2014learning} across generative~\cite{naiman2024generative, naiman2024utilizing, ren2024learning} and augmentation~\cite{kaufman2024first, nochumsohn2024beyond} tasks, among many others. Alas, not until recently, pure deep models were believed to be unable to outperform non-deep or hybrid tools~\cite{oreshkin2020nbeats}. Nevertheless, within a span of two years, pure deep methods based on RNNs~\cite{salinas2020deepar}, feedforward networks~\cite{oreshkin2020nbeats}, and the Transformer~\cite{zhou2021informer} have appeared, demonstrating competitive results and setting a new SOTA bar for long-term TSF.

Following these breakthroughs, numerous works have emerged, most of them are based on the Transformer architecture \cite{wu2021autoformer, zhou2022fedformer, zhang2022crossformer}. Recently, a surprising work~\cite{zeng2023Transformers} has shown remarkable TSF results with a simple single-layer linear model, competing and even surpassing the best models for that time. Notably, the linear model applied weights along the time axis, in contrast to the more conventional practice of applying weights along the variate axis. Subsequently, new SOTA techniques \cite{nie2023time, xue2023make} and a recent foundation model for time series \cite{zhou2023one} have integrated a similar linear module as their final decoder to attain better forecasts.


\paragraph{Multi-task learning.} Improving learning on multiple tasks follows three different categories: \emph{gradient manipulation}, \emph{task grouping}, and \emph{architecture design}. Manipulating gradients is done by imposing weights~\cite{groenendijk2020multi}, shift gradient directions~\cite{yu2020gradient}, and add an optimization step to balance gradients or resolve conflicts~\cite{chen2018gradnorm, sener2018multi, liu2021conflict}. In task grouping, previous works attempted to cluster different tasks based on similarity measures, thereafter assigning them to separate models~\cite{zamir2018taskonomy, standley2020tasks, fifty2021efficiently, song2022efficient}. Lastly, multiple forms of architecture design have been introduced to support MTL including hard-parameter sharing~\cite{vandenhende2021multi}, soft-parameter sharing~\cite{misra2016cross, ishihara2021multi}, and mixing solutions~\cite{guangyuan2022recon}. For time series data, a multi-gate mixture of experts for classification was suggested in~\cite{ma2018modeling}, whereas a soft-parameter sharing RNN-based model was proposed in~\cite{chen2020multi}. However, most of the  multi-task techniques mentioned above focus mostly on vision problems. Consequently, little attention has been given to analyzing and mitigating the multi-task learning challenges for time series problems, particularly TSF.

\section{METHOD}
\label{sec:method}

\begin{figure*}[!t]
  \centering
  \includegraphics[width=1.0\linewidth]{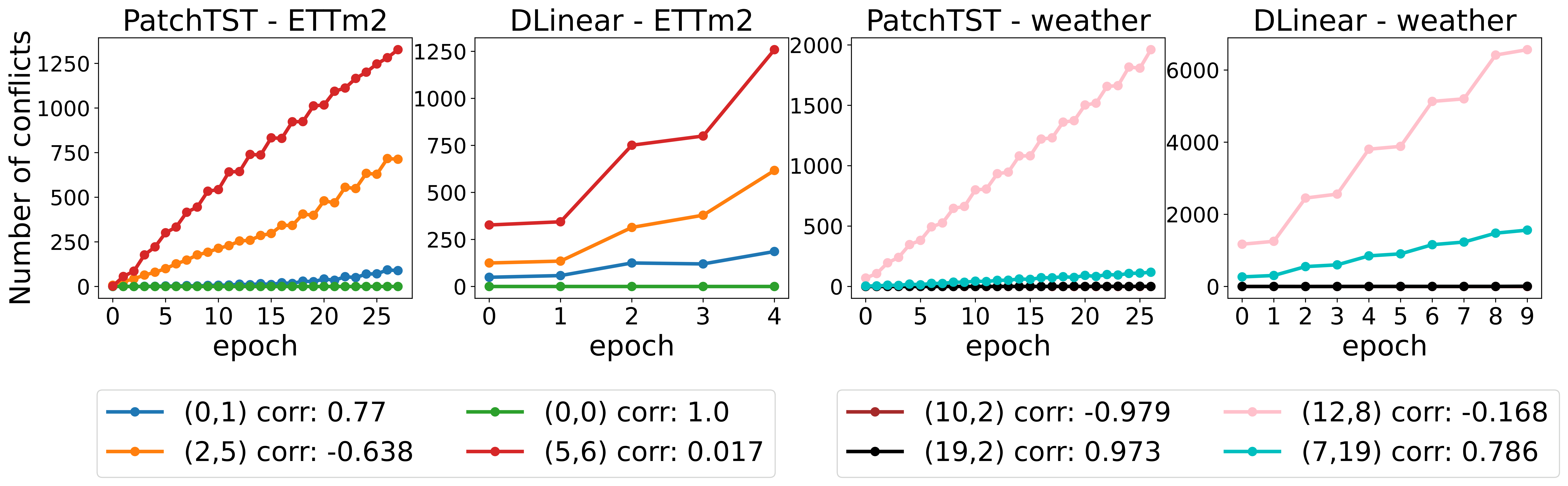}
  \vspace{-5mm}
  \caption{The total number of conflicts as a function of epochs. Colored lines represent variate pairs. Pairs with a higher absolute correlation (shown in legend) tend to have fewer conflicts during training. }
  \label{fig:conflicts}
\end{figure*}

We analyze linear forecasting (Sec.~\ref{subsec:grad_analysis}), motivating our variate grouping (Sec.~\ref{subsec:variate_grouping}), gradient scaling~\ref{subsec:variate_scaling}, and yielding an efficient and effective TSF model (Sec.~\ref{subsec:mtlinear}).

The following preliminary example motivates our perspective of TSF as MTL. We compare gradient conflicts vs. variate correlation by training a vanilla TSF model, and counting the number of conflicts between gradients after every epoch. We used the DLinear~\cite{zeng2023Transformers} and PatchTST~\cite{nie2023time} TSF models and ETTm2 and Weather datasets. The results are shown in Fig.~\ref{fig:conflicts}, where every line plot shows the conflicts number for a pair of variates. The correlation (corr) per pair is detailed in the legend, where $|\text{corr}| \approx 1$ and $|\text{corr}| \approx 0$ denote strong and weak linear correlations, respectively. Clearly, variates with a strong linear relationship (negative or positive) have fewer conflicts and vice versa, consistent across both DLinear and PatchTST architectures.

\subsection{Linear Analysis}
\label{subsec:grad_analysis}

To motivate our approach, we present below a gradient analysis of the linear model in Eq.~\eqref{eq:linear_tsf}. We begin by noting that since $\Theta$ is applied in the temporal domain, every forecast horizon $j=1,\dots,h$ is obtained with a separate set of weights, i.e., $\theta_j \in \mathbb{R}^l$ and $b_j \in \mathbb{R}$. Indeed, we have that $X^T \Theta = [X^T \theta_1, X^T \theta_2, \dots, X^T \theta_h] = [\tilde{Y}_1,\tilde{Y}_2,\dots,\tilde{Y}_h]$, where $\tilde{Y}_j$ represents the $j$-th row of $\tilde{Y}^T$, and we incorporated the bias $b_j$ in $\theta_j$, for all $j$. Thus, the MSE loss corresponding to Eq.~\eqref{eq:linear_tsf} is given by
\begin{equation} \label{eq:linear_loss}
    F(\Theta) = |X^T \Theta - Y^T|_F^2  := \sum_{j=1}^h \sum_{i=1}^k \frac{(x_i^T \theta_j - y_{j, i})^2}{kh} \ , 
\end{equation}
where $Y \in \mathbb{R}^{h \times k}$ are the true values, and we split every $X^T \theta_j$ to individual coordinates via $x_i^T \theta_j$. The objective in Eq.~\eqref{eq:linear_loss} is quadratic, and thus, its gradient is linear in $\Theta$. Given the above discussion, $F$ can be viewed either as a function $F(\Theta):\mathbb{R}^{(l+1) \times h} \rightarrow \mathbb{R}$ or as a sum of functions $F(\theta_j):\mathbb{R}^{l+1} \rightarrow \mathbb{R}$ for $j=1,\dots,h$. The gradient for the latter form reads
\begin{equation} \label{eq:linear_grad}
    \nabla_{\theta_j} F(\theta_j) = \frac{1}{k} \sum_{i=1}^k 2 x_i(x_i^T \theta_j - y_{j, i}) \ .
\end{equation}
We provide the full derivation in App.~\ref{app:grad_prod}. The expression in Eq.~\eqref{eq:linear_grad} highlights that multivariate TSF and MTL are closely related, in the sense that the $k$ variates represent $k$ different tasks sharing the same set of weights $\theta_j$. Thus, multivariate forecasting may potentially face similar challenges as multi-task learning.

The simplicity of the linear model and its gradients allows to characterize the direction and scale of gradients. Based on Eq.~\eqref{eq:linear_grad}, we arrive at the following two straightforward observations: 1) The \emph{direction} of the gradient is governed by $x_i \in \mathbb{R}^{l+1}$; and 2) The gradient's \emph{scale} is affected by $2(x_i^T \theta_j - y_{j, i}) \in \mathbb{R}$. Using these observations, we also consider the relation of different tasks, i.e., the angle and balance between the gradients associated with different variates $x_a$ and $x_b$.

To this end, the first observation reveals that each task is updated simply along the direction of its corresponding variate $x_i$. Thus, the \emph{angle} between gradients is equivalent to the angle $\alpha$ between variates $x_a$ and $x_b$, which is given by their cosine similarity, $\cos_{x_a, x_b}(\alpha) = x_a^T x_b / (| x_a | \cdot | x_b |)$.
The cosine similarity is closely related to the \emph{Pearson Correlation Coefficient} (PCC)~\cite{cohen2009pearson}. Indeed, PCC and cosine similarity coincide for centered data. Thus, we interchangeably use these terms below. In our work, we consider high absolute similarity, i.e., $|\cos_{x_a, x_b}(\alpha)| \approx 1$, to correspond to a strong relationship between tasks, whereas low PCC, $|\cos_{x_a, x_b}(\alpha)| \approx 0$, means weakly-related variates. 

The second observation details what impacts the gradient norm. Suppose some variates are inherently more difficult to forecast than others due to rapid distribution shifts with the lack of clear seasonality or trend, thus achieving a larger error $e_{i,j} = |x_i^T\theta_j - y_{j,i}|$. Clearly, in such cases the gradient will dominate the optimization process due to its large magnitude. In other words, gradients associated with variates with higher forecast uncertainty have generally larger gradients, see Fig.~\ref{fig:grad_mag}. Based on the above analysis and corresponding observations and towards addressing multi-task learning issues related to the tragic triad, we propose in what follows a simple, efficient and effective method that takes into account the correlation of variates and the error $e_{i,j}$.



\subsection{Variate grouping}
\label{subsec:variate_grouping}

Based on Sec.~\ref{sec:related}, one common approach to deal with some of the challenges underlying MTL is to assign non-conflicting tasks to separate sets of weights~\cite{fifty2021efficiently, guangyuan2022recon, song2022efficient}. A direct implementation of this idea in TSF will result in an independent model per variate, which may be highly demanding computationally for data with many variates (App.~\ref{app:mem}). An optional middle-ground toward an efficient framework is to assign separate weights for \emph{a-priori clustered variates}. Namely, we identify similar variates, clustering them into groups with separate network weights. Importantly, forecasting is different from multi-task learning problems in, e.g., vision, where task relationships can not be necessarily extracted directly from the input. In comparison, the objective in TSF is to accurately predict the future horizon, which shares fundamental features with the input lookback, such as trend, seasonality, periodicity, and data distribution. Moreover, our analysis in Sec.~\ref{subsec:grad_analysis} shows that tasks' alignment can be deduced in practice by estimating the correlation between different variate vectors. 

Our a-priori clustering and weights assignment method follows three simple steps: 1) Compute the absolute-value correlation matrix for variates, defined via $R_X = \left( | \cos_{x_a, x_b}(\alpha)| \right)$ for $a,b=1,\dots, k$.
2) Perform agglomerative clustering on $R_X$, based on a threshold $\bar{\alpha}$, encoding the maximum angle between two variates. Note that variates with $|\alpha| < \bar{\alpha}$ are grouped together. Sec.~\ref{subsec:abl_lookback_grouping} justifies empirically grouping variates with a strong negative correlation, as they share a similar optimization trajectory. 3) Finally, for each variates cluster, assign a separate linear-based neural network (Sec.~\ref{subsec:mtlinear}).  

\begin{figure*}
    \centering
    \begin{overpic}[width=1.0\linewidth]{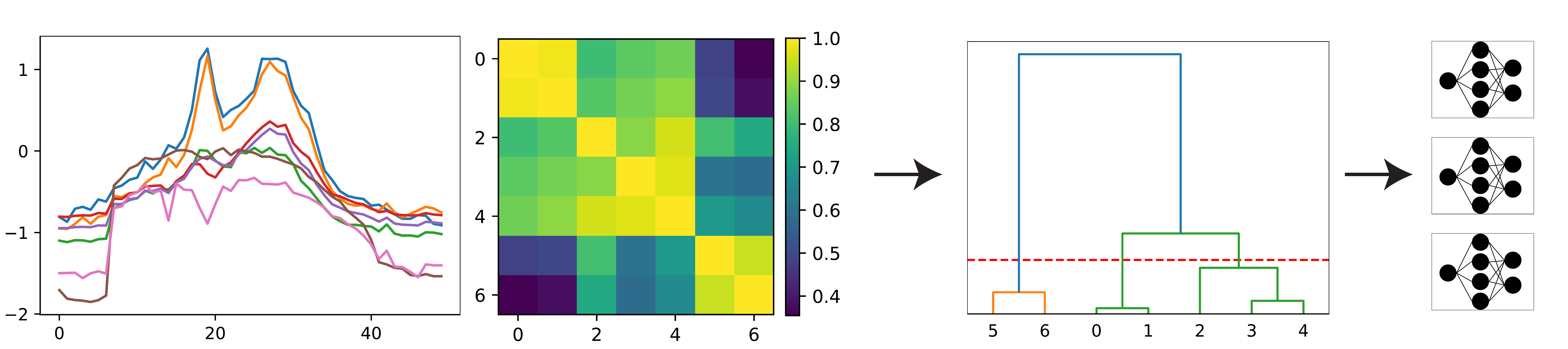}
        \put(28, 21){Step 1} \put(70, 21){Step 2} \put(92, 21){Step 3}
    \end{overpic}
    \vspace{-5mm}
    \caption{Our pipeline consists of three steps: estimating variate correlations, variate clustering, and assigning a linear module per group. The resulting framework, \textbf{MTLinear} solves multivariate TSF effectively.}
    \label{fig:mtlinear}
\end{figure*}

\begin{table*}[ht]
    \caption{Multivariate forecasting results of MTLinear (ours) compared to other strong baselines. A \textbf{bold} and \underline{underlined} notation represent the best and second-best scores, respectively. }
    \centering
    \resizebox{\linewidth}{!}{
\begin{tabular}{l|rr|rr|rr|rr|rr|rr|rr|rr|}
\toprule
        &  \multicolumn{2}{c|}{MTDLinear} & \multicolumn{2}{c|}{MTNLinear} &  \multicolumn{2}{c|}{iTransformer} &  \multicolumn{2}{c|}{PatchTST} &  \multicolumn{2}{c|}{Crossformer} &  \multicolumn{2}{c|}{DLinear} & \multicolumn{2}{c|}{FEDformer} &  \multicolumn{2}{c|}{Autoformer} \\
\multicolumn{1}{c|}{Dataset}  &       MSE&MAE                  &                 MSE&MAE                    &                      MSE&MAE                         &                 MSE&MAE                     &                     MSE&MAE                        &                 MSE&MAE                    &                   MSE&MAE                      &                    MSE&MAE                       \\
\midrule
ETTm1    &            \underline{0.399} &            \underline{0.402} &            0.403 &            \underline{0.402} &                0.407 &                0.410 &           \textbf{ 0.387} &            \textbf{ 0.400} &               0.513 &               0.495 &           0.403 &           0.407 &             0.448 &             0.452 &              0.588 &              0.517 \\
ETTm2    &            0.284 &            0.334 &            \textbf{ 0.279} &            \textbf{ 0.320} &                0.288 &                0.332 &            \underline{0.281} &            \underline{0.326} &               0.757 &               0.610 &           0.350 &           0.401 &             0.304 &             0.349 &              0.327 &              0.371 \\
ETTh1    &            0.456 &            0.441 &            \underline{0.444} &           \textbf{  0.429} &                0.454 &                0.448 &            0.469 &            0.454 &               0.529 &               0.522 &           0.456 &           0.452 &            \textbf{0.440} &             0.460 &              0.496 &              0.487 \\
ETTh2    &            0.453 &            0.447 &            \textbf{ 0.373} &            \textbf{ 0.397} &                \underline{0.383} &                \underline{0.406} &            0.387 &            0.407 &               0.942 &               0.684 &           0.559 &           0.515 &             0.436 &             0.449 &              0.450 &              0.459 \\
ECL      &            \underline{0.198} &            0.286 &            0.204 &            \underline{0.283} &                \textbf{ 0.178} &                \textbf{ 0.270} &            0.205 &            0.290 &               0.244 &               0.334 &           0.212 &           0.300 &             0.214 &             0.327 &              0.227 &              0.338 \\
Exchange &            \textbf{ 0.290} &            \textbf{ 0.377} &            0.410 &            0.422 &                0.360 &                \underline{0.403} &            0.366 &            0.404 &               0.940 &               0.707 &           \underline{0.354} &           0.414 &             0.518 &             0.429 &              0.613 &              0.539 \\
Traffic  &            0.621 &            0.380 &            0.624 &            0.372 &                \textbf{ 0.428} &                \textbf{ 0.282} &            \underline{0.481} &            \underline{0.304} &               0.550 &               \underline{0.304} &           0.624 &           0.383 &             0.609 &             0.376 &              0.628 &              0.379 \\
Weather  &            \textbf{ 0.238} &             0.295&            \underline{0.249} &            \textbf{ 0.276} &                0.258 &                \underline{0.278} &            0.258 &            0.280 &               0.258 &               0.315 &           0.265 &           0.317 &             0.309 &             0.360 &              0.338 &              0.382 \\
ILI      &           \underline{ 2.234} &            \underline{0.995} &            \textbf{ 1.964} &            \textbf{ 0.902} &                2.738 &                1.098 &            2.421 &            1.011 &               3.386 &               1.236 &           2.616 &           1.090 &             2.846 &             1.144 &              3.006 &              1.161 \\
\hline
 Average & \underline{0.575} &            0.440 &            \textbf{0.550} &            \textbf{0.423} &                0.610 &                0.436 &            0.584 &            \underline{0.431} &               0.902 &               0.579 &           0.649 &           0.475 &             0.680 &             0.483 &              0.741 &              0.515 \\
$1^{st}$ Count &            \underline{2} &            1 &            \textbf{3} &            \textbf{5} &                \underline{2} &                \underline{2} &            1 &            1 &               0 &               0 &           0 &           0 &             1 &             0 &              0 &      0 \\
\bottomrule
\end{tabular}}
\label{tab:main_results_96}
\end{table*}

\subsection{Gradient manipulation}
\label{subsec:variate_scaling}

Even though dissimilar variates are assigned to different sets of weights, the risk of having a subset of variates dominating the optimization process within each group still prevails. Thus, we introduce a \emph{gradient magnitude penalty} $w_{i,j}^a \in \mathbb{R}^+$ that incorporates the error $e_{i,j}$ of its associated gradient. The penalty $w_{i,j}^a$ multiplies the loss for every admissible $i, j$, and the new loss is formally given by $F_W(\Theta) = | X^T \Theta - Y^T|_{W^a}^2$, where
\begin{equation} \label{eq:our_loss}
    | X^T \Theta - Y^T|_{W^a}^2 = \sum_j^h \sum_i^k  w_{i,j}^a(\mathbf{x}_i^T\theta_j - y_{i,j})^2 \ , 
\end{equation}
where $|A|^2_{W^a} = \text{trace} [A^T \odot \sqrt{W^a} (\sqrt{W^a})^T \odot A]$, $W^a = (w_{i,j}^a) \in \mathbb{R}^{k \times h}$, and the weights $w_{i, j}^a$ are defined as follows,
\begin{equation} \label{eq:our_scaling}
    w_{i,j}^a = \frac{1}{(K_j \cdot H_i)^a} , \ K_j = \sum_{i=1}^k \frac{e_{i,j}}{k}  , \  H_i = \sum_{j=1}^h \frac{e_{i,j}}{h} .
\end{equation}

Essentially, $w_{i,j}^a$ addresses dominant scales along the horizon axis and the variates axis. Specifically, $K_j$ is the mean error of different variates for the same horizon $j$, whereas $H_i$ is the mean error of different horizons for a given variate $i$. Thus, $w_{i,j}^a$ balances both means when they attain high magnitudes. The parameter $a$ controls the intensity of our penalty. During training, we treat $w_{i,j}^a$ as a constant scalar (i.e., a computational graph leaf), thus Eq.~\eqref{eq:our_loss} avoids additional gradient computations. Consequently, the computational complexity of our gradient manipulation is $\mathcal{O}(1)$, unlike other methods~\cite{chen2018gradnorm, yu2020gradient, liu2021conflict} whose complexity is $\mathcal{O}(k)$. This procedure is applied individually to each group's linear model.


\subsection{Multi-task Linear Model}
\label{subsec:mtlinear}

The individual linear models mentioned in Sec.~\ref{subsec:variate_grouping} are added to a training framework we call \textbf{MTLinear}. MTLinear is a single multi-head linear layer with $c$ heads, each corresponding to a group of variates. Importantly, the separate heads can be trained in parallel, reducing the overall computational footprint of our approach. Each head of MTLinear is based on the linear models proposed in~\cite{zeng2023Transformers}. Specifically, we focus on the DLinear and NLinear baselines. DLinear decomposes the time series to a trend component and a remainder component, where each component is handled by a separate set of weights. The trend is extracted with a standard average pooling kernel. NLinear introduces a ``normalization'' pre-processing mechanism where the last values of the lookback are subtracted from the series before the forward pass, and are added back in when computation finishes. Our overall pipeline for time series forecasting is illustrated in Fig.~\ref{fig:mtlinear}.


\begin{figure*}[t]
  \centering
  \includegraphics[width=1.0\linewidth]{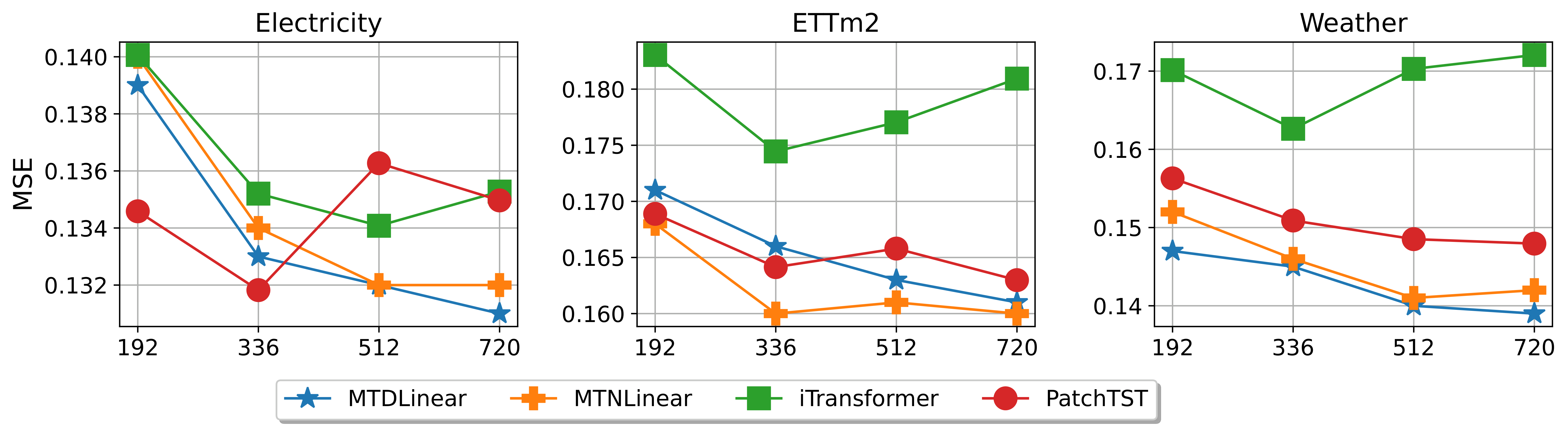}
  \vspace{-5mm}
  \caption{MSE results for different lookback lengths with a forecast horizon of $96$.}
  \label{fig:ablate_lookback}
\end{figure*}

\begin{table*}[t]
    \caption{Average MSE scores ablating the components of MTLinear. See text.}
    \label{tab:breakdown}

    \begin{minipage}[c]{0.5\textwidth}
    \centering
    \scalebox{0.8}{
    \begin{tabular}{|l|c|c|c|c|}
        \toprule
        & \multicolumn{3}{|c|}{MTDLinear}& \\ 
        Dataset &   Ours &  Penalty &  Grouping &  Baseline \\
        \midrule
        ETTm1    &  \textbf{0.399} &    0.402 &     \textbf{0.399} &     0.406 \\
        ETTm2    &  \textbf{0.284} &    \textbf{0.284} &     0.323 &     0.323 \\
        ETTh1    &  \textbf{0.456} &    0.464 &     \textbf{0.456} &     0.471 \\
        ETTh2    &  \textbf{0.453} &    \textbf{0.453} &     0.503 &     0.498 \\
        ECL      &  \textbf{0.198} &    0.209 &     0.199 &     0.209 \\
        Exchange &  0.289 &    \textbf{0.278} &     0.312 &     0.309 \\
        Traffic  &  \textbf{0.621} &    0.626 &     0.622 &     0.625 \\
        Weather  &  \textbf{0.238} &    0.264 &     0.239 &     0.267 \\
        ILI      &  \textbf{2.234} &    \textbf{2.234} &     2.769 &     2.728 \\
        \bottomrule
    \end{tabular}}
    \end{minipage}
    \begin{minipage}[c]{0.5\textwidth}
    \centering
    \scalebox{0.8}{
    \begin{tabular}{|l|c|c|c|c|}
        \toprule
        & \multicolumn{3}{|c|}{MTNLinear} &  \\ 
        Dataset & Ours & Penalty & Grouping & Baseline \\ 
            \midrule
        ETTm1    &  \textbf{0.403} &    0.407 &     0.404 &     0.410 \\
        ETTm2    &  \textbf{0.280} &    0.285 &     0.281 &     0.286 \\
        ETTh1    &  0.443 &    \textbf{0.442} &     0.444 &     0.446 \\
        ETTh2    &  \textbf{0.373} &    0.374 &     0.374 &     0.374 \\
        ECL      &  \textbf{0.204} &    0.215 &     \textbf{0.204} &     0.214 \\
        Exchange &  0.410 &    \textbf{0.366} &     0.411 &     0.378 \\
        Traffic  &  \textbf{0.624} &    0.625 &     \textbf{0.624} &     \textbf{0.624} \\
        Weather  &  \textbf{0.249} &    0.272 &     \textbf{0.249} &     0.273 \\
        ILI      &  \textbf{1.965} &    \textbf{1.965} &     2.254 &     2.213 \\
            \bottomrule
    \end{tabular}}
    \end{minipage}
\end{table*}

\section{EXPERIMENTS}
\label{sec:exp}

Details regarding the datasets, models, experimental setup and implementation notes are provided in App.~\ref{app:eval_setup}. We tested our approach based on DLinear and NLinear vs. SOTA nonlinear Transformer models. The results in Tab.~\ref{tab:main_results_96} show the MSE and MAE scores. Each score represents the average of four forecast horizons $h \in \{24, 36, 48, 60\}$ and 36 input length for ILI, as well as  $h \in \{96, 192, 336, 720\}$ and 96 input length for the remaining. We use a similar format also in Tabs.~\ref{tab:breakdown}, \ref{tab:grad_manip}, and \ref{tab:linear_compare}. Our results indicate that MTLinear has an overall superior performance with a best global MSE and MAE corresponding to \textbf{0.550} and \textbf{0.423} with MTNLinear, and second best MSE \textbf{0.575} with MTDLinear. In general, we outperform the transformer models, obtaining 11 top scores whereas iTransformer and PatchTST account for only 6. The full results are presented in Tab.~\ref{tab:main_results_96_full}, along with an extension to the 336 lookback length where we also compare to the original PatchTST and GPT4TS results reported in Tab.~\ref{tab:main_results_336_full}.

\subsection{Ablation of MTLinear Components}

The proposed MTLinear method incorporates variate clustering and a gradient penalty, which, although separate components, work synergistically to address the tragic triad issues. Specifically, clustering helps manage gradient conflicts, while scaling addresses the varying gradient magnitudes. In the following experiment, we evaluate the effect of each component individually and in combination, with results presented in Tab.~\ref{tab:breakdown}. Our ablation study reveals that in some datasets, using only one component may provide minimal benefit or even degrade performance. However, the combined approach consistently outperforms, with both components together yielding superior results in most cases.

\subsection{Longer Lookbacks and Grouping Criteria}
\label{subsec:abl_lookback_grouping}

\paragraph{Longer lookbacks.} Constraining a forecast model to a fixed lookback, such as 96, reduces its robustness and adaptability to different problems with varying lookback lengths. Here, we evaluate MTLinear’s performance across different lookbacks—192, 336, 512, and 720—and compare it with iTransformer and PatchTST. As shown in Fig.~\ref{fig:ablate_lookback}, MTLinear variants consistently outperform iTransformer across all cases, with particularly strong results for the 720 lookback. While PatchTST proves competitive, MTLinear surpasses its performance across all lookbacks and datasets, except for Electricity at 192 and 336 lookbacks. Overall, MTLinear consistently improves as the lookback length increases, a trend not typically observed with transformer-based models~\cite{zeng2023Transformers}.

\begin{figure*}[t]
  \centering
  \includegraphics[width=1.0\linewidth]{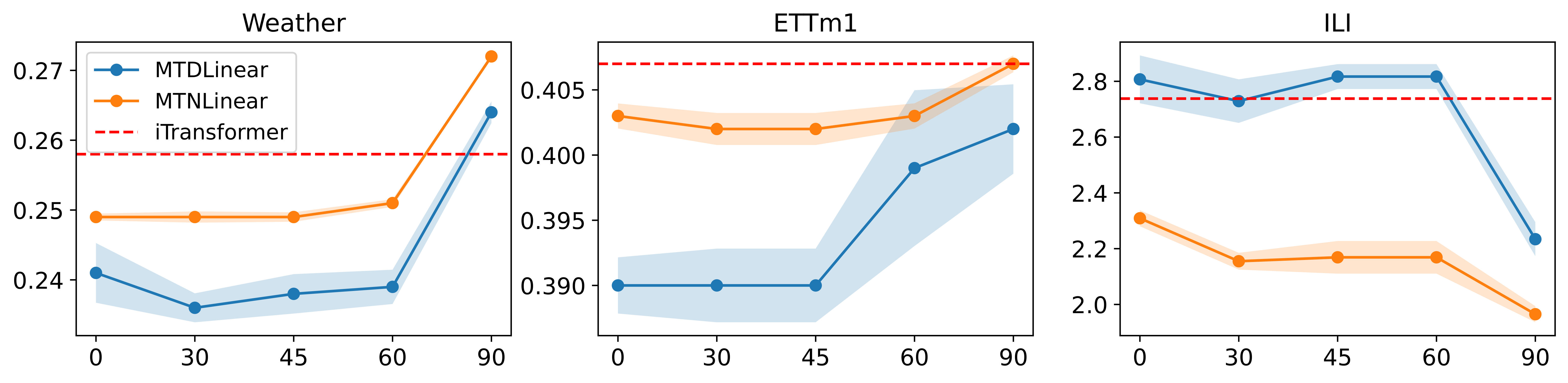}
  \vspace{-5mm}
  \caption{MSE measures for different clustering $\bar{\alpha}$. The red dashed line is the mean for iTransformer. The results suggest that MTLinear is comparable or better in comparison to iTransformer.}
  \label{fig:phi_analysis}
\end{figure*}

\begin{table*}[!t]
    \caption{A comparison of gradient manipulation techniques to MTLinear with DLinear NLinear.}
    \begin{minipage}[c]{1.0\textwidth}
    \centering
    \scalebox{0.79}{
    \begin{tabular}{|l|c|c|c|c|c|c|c|}
        \toprule
        & \multicolumn{7}{|c|}{DLinear} \\ 
        Dataset & Ours & GradNorm & CoV-W  & PCgrad & CAGrad &  Nash-MTL & Baseline \\ 
        \midrule
        ETTm1 & \textbf{0.399} & 0.472 & 0.406 & 0.411& 0.406 & 0.406 & 0.406 \\ 
        ETTm2 & \textbf{0.284} & 0.447 & 0.316 & 0.302 & 0.306 & 0.325 & 0.323\\
        Exchange & \textbf{0.289} & 0.379 & 0.311 & 0.325 & 0.375 & 0.375 & 0.309\\
        Weather & \textbf{0.244} & 0.280 & 0.268 & 0.266 & 0.267 & 0.267 & 0.267\\ 
        ILI & \textbf{2.234} & 4.850 & 2.600 & 2.443 & 2.350 & 2.780 & 2.728\\ 
    \bottomrule
    \end{tabular}}
    \end{minipage}
    
    \vspace*{0.1 cm}
    
    \begin{minipage}[c]{1.0\textwidth}
    \centering
    \scalebox{0.79}{
    \begin{tabular}{|l|c|c|c|c|c|c|c|}
        \toprule
         & \multicolumn{7}{|c|}{NLinear} \\ 
        Dataset & Ours & GradNorm & CoV-W  & PCgrad & CAGrad & Nash-MTL & Baseline \\ 
        \midrule
        ETTm1 & \textbf{0.403} & 0.438 & 0.410 & 0.414 & 0.414 & 0.413 & 0.410 \\ 
        ETTm2 & \textbf{0.280} & 0.299 & 0.285	 & 0.284 & 0.284 & 0.286 & 0.286 \\
        Exchange & 0.410 & 0.419 & 0.379 & 0.371 & \textbf{0.357} & \textbf{0.357} & 0.378\\
        Weather & \textbf{0.249} & 0.306 & 0.274 & 0.273 & 0.273 & 0.274 & 0.273\\
        ILI & \textbf{1.965} & 2.540 & 2.174 & 2.150 & 1.987 & 2.201 & 2.213\\ 
 \bottomrule
    \end{tabular}}
    \label{tab:grad_manip}
\end{minipage}
\end{table*}

\begin{table*}[h!]
\caption{MTLinear based on different linear modules.}
    \centering
    \resizebox{\linewidth}{!}{
\begin{tabular}{l|rrl||rrl||rrl||rrl|}
\toprule
{} &  \multicolumn{3}{c||}{MTLinear} &  \multicolumn{3}{c||}{MTRLinear} &  \multicolumn{3}{c||}{MTDLinear} &  \multicolumn{3}{c|}{MTNLinear} \\
Dataset &   Ours &  Baseline &   \% Imp &   Ours &  Baseline &  \% Imp &   Ours &  Baseline &     \% Imp &   Ours &  Baseline &    \% Imp  \\
\midrule
ETTm1    &       \textbf{0.397} &            0.413 &   3.87\% &        \textbf{0.400} &             0.413 &   3.15\% &        \textbf{0.399} &             0.406 &   1.72\% &        \textbf{0.403} &             0.410 &   1.71\% \\
ETTm2    &       \textbf{0.291} &            0.324 &  10.19\% &        \textbf{0.280} &             0.286 &    2.1\% &        \textbf{0.284} &             0.323 &  12.07\% &        \textbf{0.280} &             0.286 &    2.1\% \\
ETTh1    &       \textbf{0.456} &            0.464 &   1.72\% &        0.445 &             0.445 &    0.0\% &        \textbf{0.456} &             0.471 &   3.18\% &       \textbf{ 0.443} &             0.446 &   0.67\% \\
ETTh2    &       \textbf{0.471} &            0.489 &   3.68\% &        \textbf{0.376} &             0.377 &   0.27\% &        \textbf{0.453} &             0.498 &   9.04\% &        \textbf{0.373} &             0.374 &   0.27\% \\
ECL      &       \textbf{0.198} &            0.209 &   5.26\% &        \textbf{0.203} &             0.214 &   5.14\% &        \textbf{0.198} &             0.209 &   5.26\% &        \textbf{0.204} &             0.214 &   4.67\% \\
Exchange &       \textbf{0.284} &            0.289 &   1.73\% &        0.370 &             \textbf{0.359} &  -3.06\% &        \textbf{0.289} &             0.309 &   6.47\% &        0.410 &             \textbf{0.378} &  -8.47\% \\
Traffic  &       \textbf{0.621} &            0.625 &   0.64\% &        0.623 &             0.623 &    0.0\% &        \textbf{0.621} &             0.625 &   0.64\% &        0.624 &             0.624 &    0.0\% \\
Weather  &      \textbf{ 0.241} &            0.268 &  10.07\% &        \textbf{0.244} &             0.272 &  10.29\% &        \textbf{0.238} &             0.267 &  10.86\% &        \textbf{0.249} &             0.273 &   8.79\% \\
ILI      &       \textbf{2.320} &            2.858 &  18.82\% &        \textbf{2.148} &             2.423 &  11.35\% &        \textbf{2.234} &             2.728 &  18.11\% &        \textbf{1.965} &             2.213 &  11.21\% \\
\bottomrule
\end{tabular}}
\label{tab:linear_compare}
\end{table*}

\paragraph{Grouping criteria.} We now examine the effect of various angles, $\bar{\alpha} \in {\pi/2, \pi/3, \pi/4, \pi/6, 0}$, each representing a maximum cosine similarity threshold for grouping variates. Specifically, $\bar{\alpha} = 0$ enforces strict clustering, with one variate per cluster, while $\bar{\alpha} = \pi/2$ results in a single model across all variates, as groups are formed based on absolute correlation values. In Fig.~\ref{fig:phi_analysis}, we plot MTLinear’s performance for each $\bar{\alpha}$ on the $x$-axis, with corresponding MSE scores on the $y$-axis. Our findings indicate that $\bar{\alpha}$ impacts datasets differently, likely due to varying degrees of the tragic triad issues. Weather and ETTm1 perform well with most groupings, suggesting that a shared model does not eliminate gradient conflicts. Conversely, for ILI, $\bar{\alpha} = \pi/2$ proves optimal. Additionally, model size plays a role, as groupings with $\bar{\alpha} > 0$ yield a more compact model—especially relevant for datasets with numerous variates. Tab.~\ref{tab:number_groups} shows that this approach can reduce model size by more than half while enhancing performance. Notably, in each sub-plot of Fig.~\ref{fig:phi_analysis}, several MTLinear configurations outperform iTransformer, indicating that linear forecasters remain effective in TSF and merit inclusion in new methods. Finally, we study the correlation-conflict similarity in Fig.~\ref{fig:conflict_matrix}, finding that the correlation and conflict matrices are similar for DLinear and PatchTST but not for Autoformer. This difference may stem from PatchTST and DLinear’s use of weight-per-time linear layers, unlike Autoformer.

\begin{figure*}[h]
  \centering
  \includegraphics[width=1.0\linewidth]{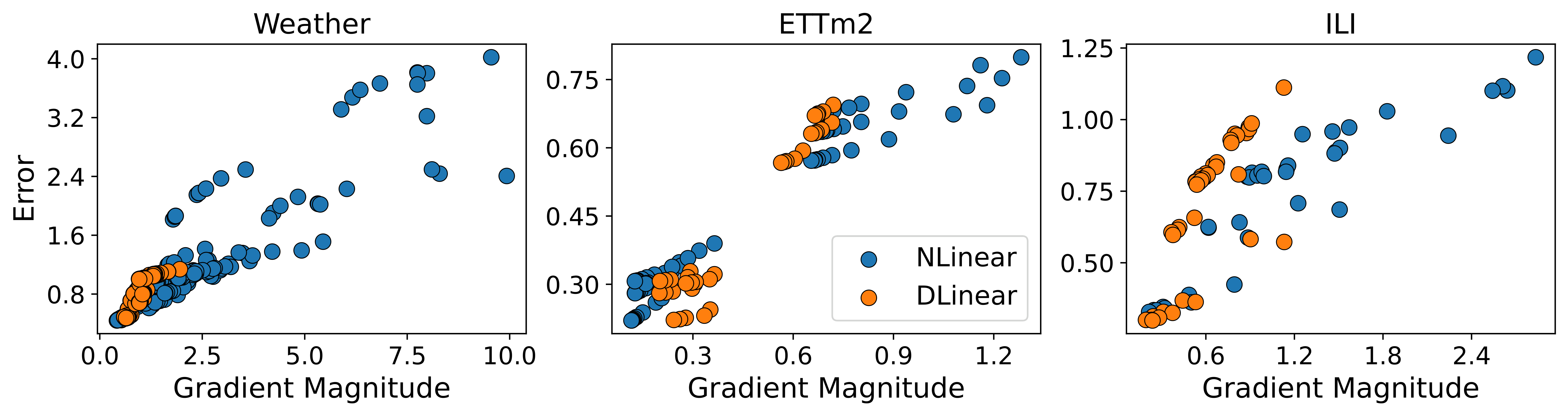}
  \vspace{-5mm}
  \caption{We plot the error $e_{i,j}$ of a given loss vs. its gradient's magnitude. These results highlight the clear positive correlation between the two for both DLinear and NLinear.}
  \label{fig:grad_mag}
\end{figure*}

\subsection{Variants of Gradient Manipulation}

We compare our approach with robust gradient manipulation baselines, including GradNorm~\cite{chen2018gradnorm} and PCGrad~\cite{yu2020gradient}, which adjust gradients per task to address magnitude and conflict issues. Additionally, we benchmark against Cov-Weighting (CoV-W)~\cite{groenendijk2020multi}, which emphasizes tasks with higher trailing variance, assuming that reduced variance indicates a satisfied loss. We also include CAGrad~\cite{liu2021conflict} and Nash-MTL~\cite{navon2022multi}. Exploring different scaling schemes helps assess whether the variance of $e_{i,j}$ impacts results beyond the error alone. As shown in Tab.~\ref{tab:grad_manip}, MTLinear outperforms these methods while being more efficient, requiring only a single backward pass compared to the $k$ passes needed by PCGrad and GradNorm. We further analyze the link between error term $e_{i,j}$ and gradient magnitude in Fig.~\ref{fig:grad_mag}, where we find a strong correlation across the Weather, ETTm2, and ILI datasets, with higher errors corresponding to larger gradients—motivating our penalty in Sec.~\ref{subsec:variate_scaling}.

\begin{table}[b!]
    \caption{Linear probing with MTLinear (MTLi) using PatchTST on source datasets Electricity (top) and Weather (bottom).
    The model is then tested on target datasets. See text for details.}
\begin{minipage}[c]{0.5\textwidth}

    \centering
    \scalebox{0.75}{
    \begin{tabular}{|l|c|c|c|c|}
        \toprule
        Dataset & MTLi & MTLi Probing & Zero-shot & PatchTST \\ 
        \midrule
        ECL* & $\mathbf{0.161}$ & $\mathbf{0.161}$ & $0.162$ & $0.162$ \\ 
        \midrule
        Weather & $\mathbf{0.22}$ & $0.226$ & $0.326$ & $0.229$ \\ 
        ETTm2 & $\mathbf{0.254}$ & $0.258$ & $0.344$ & $0.257$ \\ 
        \bottomrule
    \end{tabular}}
    \end{minipage}
    \begin{minipage}[c]{0.5\textwidth}
    \centering
    \scalebox{0.75}{
    \begin{tabular}{|l|c|c|c|c|}
        \toprule
        Dataset & MTLi & MTLi Probing & Zero-shot & PatchTST \\ 
        \midrule
        Weather* & $\mathbf{0.22}$ & $0.224$ & $0.231$ & $0.231$ \\ 
        \midrule
        ECL & $\mathbf{0.161}$ & $0.165$ & $0.867$ & $0.162$ \\ 

        ETTm2 & $\mathbf{0.254}$ & $\mathbf{0.254}$ & $0.313$ & $0.257$ \\  \bottomrule
    \end{tabular}}

    \label{tab:linear_prob}
\end{minipage}

\end{table}

\subsection{Model Variations and Linear Probing}

The MTLinear method is versatile and can integrate with linear layers beyond NLinear or DLinear, including a standard linear layer, to boost performance. We present these results in Tab.~\ref{tab:linear_compare}, where MTLinear-based approaches consistently improve baseline MSE scores by up to 18\% for ILI and 10\% across other datasets.


In TSF transfer learning with foundation models and Transformer architectures, a common approach is linear probing, where only the final linear layer is fine-tuned on a pre-trained model~\cite{nie2023time, zhou2023one, jin2023time}. To assess the transfer capabilities of TSF models, we conducted an experiment where several baselines were first trained on a source dataset and then fine-tuned on a target dataset. The baselines include MTLinear, MTLinear Probing, Zero-shot, and PatchTST. MTLinear Probing refers to PatchTST with our MTLinear module replacing its decoder, while Zero-shot involves no fine-tuning. MSE results are presented in Tab.~\ref{tab:linear_prob}, with the top table showing results for Electricity as the source dataset with Weather and ETTm2 as targets, and the bottom table displaying Weather as the source dataset with Electricity and ETTm2 as targets. All experiments used a 336 lookback, as in PatchTST’s original setup.

Our conclusions are as follows: 1) MTLinear consistently outperforms all transfer learning configurations, including PatchTST, on the evaluated datasets; 2) Training PatchTST with MTLinear probing further enhances results, evidenced by MSE reductions for both Electricity and Weather; and 3) MTLinear probing applied to non-pre-trained PatchTST yields significant improvements, bringing linear fine-tuning results close to, and in some cases surpassing, fully trained PatchTST outcomes. However, this configuration still does not match the performance of a standard MTLinear setup without PatchTST.

\section{CONCLUSION}

In this work, we considered the task of multivariate time series forecasting. While several strong existing works treat different variates independently, we offered to exploit their inter-relations. We do so by viewing multivariate forecasting as a multi-task learning problem, allowing to consider forecasting through the tragic triad challenges. Our analysis of linear models and their gradients suggest that variates are optimized along the direction of the variate, and scaled proportionally to the prediction error. Based on our analysis, we propose to group variates together if the angle between them is small, and to balance variate groups using their error. Ultimately, each variate group is viewed as an independent task, solved using a single linear module, and combined into an optimization framework we named MTLinear. Our approach shows competitive results in several benchmarks in comparison to strong baseline methods. While our method effectively utilizes inter-related variate information, its structure limits the exploitation of non-correlated cross-variate information. This is significant, as different variates—regardless of their correlation—may contain unique information that could enhance the overall predictive performance. Another limitation is our reliance on a specific linear layer type, such as DLinear or NLinear. We believe that a unified framework combining these or incorporating additional layer types would further strengthen the MTLinear approach.

In the future, we plan to investigate the incorporation of the multi-head MTLinear as a decoder in state-of-the-art methodologies. Additionally, we will explore the effect of learning the clusters during training, instead of computing them as a pre-processing step. Lastly, we wish to address the limitations of our work. In general, we believe that further studying the inter-relations between variates of real-world time series data is important, and our work is a first step toward achieving that goal.

\subsubsection*{Acknowledgments}
This research was partially supported by the Lynn and William Frankel Center of the Computer Science Department, Ben-Gurion University of the Negev, an ISF grant 668/21, an ISF equipment grant, and by the Israeli Council for Higher Education (CHE) via the Data Science Research Center, Ben-Gurion University of the Negev, Israel.

\clearpage
\bibliographystyle{abbrv}
\bibliography{refs}

\clearpage
\onecolumn
\appendix

\section{Linear Analysis Proof}
\label{app:grad_prod}

Below, we provide the full derivation for the gradient provided in Eq.~\eqref{eq:linear_grad}. We recall that $F(\Theta)$ can be viewed as a function $F(\Theta):\mathbb{R}^{l+1 \times h} \rightarrow \mathbb{R}$ or as a sum of functions $F(\theta_j):\mathbb{R}^{l+1} \rightarrow \mathbb{R}$. After flattening, the gradient of the first function is an object of size $l+1 \cdot h$, where every $l+1$ elements correspond to a particular gradient for $F(\theta_j)$, $j=1,\dots,h$. Thus, it is sufficient to derive the gradient of $F(\theta_j)$, as we detail below.

\begin{align*}
    F(\Theta) &= \frac{1}{kh} \sum_{j=1}^h \sum_{i=1}^k (x_i^T \theta_j - y_{j, i})^2 \ , \quad F(\theta_j) = \frac{1}{k} \sum_{i=1}^k (x_i^T \theta_j - y_{j, i})^2  \\
    F(\theta_j + \delta \theta_j) - F(\theta_j) &= \frac{1}{k} \sum_{i=1}^k (x_i^T (\theta_j + \delta\theta_j) - y_{j,i})^2 - (x_i^T \theta_j - y_{j,i})^2 \\
    &=^* \frac{1}{k} \sum_{i=1}^k x_i^T(\theta_j + \delta\theta_j)[x_i^T (\theta_j + \delta\theta_j) - y_{j,i}] - x_i^T \theta_j (x_i^T \theta_j - y_{j,i}) \\
    &= \frac{1}{k} \sum_{i=1}^k x_i^T (\theta_j + \delta\theta_j) (x_i^T \theta_j - y_{j,i}) + x_i^T (\theta_j + \delta\theta_j)(x_i^T\delta\theta_j - y_{j,i}) - x_i^T \theta_j (x_i^T \theta_j - y_{j,i}) \\
    &=^{**} \frac{1}{k} \sum_{i=1}^k x_i^T \delta\theta_j(x_i^T \theta_j - y_{j,i}) + x_i^T \delta\theta_j(x_i^T\theta_j - y_{j,i}) \\
    &= \frac{2}{k} \sum_{i=1}^k \delta\theta_j^T x_i (x_i^T \theta_j - y_{j,i}) = \delta\theta_j^T \nabla_{\theta_j} F(\theta_j) \ ,
\end{align*}
where the starred pass is where we leave only elements that depend on $\theta_j$, and the double starred pass is due to eliminating non first-order in $\delta\theta_j$ elements.

\section{Convergence Proof}

In what follows, we provide a straightforward proof for the convexity of our optimization, detailed in Eq.~\eqref{eq:our_loss}. Then, under certain mild conditions that are satisfied by our problem, stochastic gradient descent (SGD) is guaranteed to converge~\cite{boyd2004convex}. We recall that per cluster, our loss take the form of $| X^T \Theta - Y^T|_{W^a}^2$, where $X \in \mathbb{R}^{l \times k}, Y \in \mathbb{R}^{h \times k}, \theta \in \mathbb{R}^{l \times h}$, and $W^a \in \mathbb{R}^{h \times k}$. The norm $| A |_{W^a} := \text{trace}(A^T \odot W^a A)$, where $\odot$ is an element-wise multiplication operation. To prove that Eq.~\eqref{eq:our_loss} is convex, we will show that its Hessian is a fixed, semi-positive definite (SPD) matrix. First, we observe that since each element $W^a_{ij} \geq 0$ by Eq.~\eqref{eq:our_scaling}, then it holds that $|A|^2_{W^a} = \text{trace} [A^T \odot \sqrt{W^a} (\sqrt{W^a})^T \odot A]$, where $\sqrt{A}$ is the element-wise square-root of the matrix $A$. Additionally, we denote by $A_W$ a matrix scaled by $\sqrt{W^a}$, i.e., $A_W = (\sqrt{W^a})^T \odot A$. Then, it follows that
\begin{align*}
    |X^T \Theta - Y^T|_{W^a}^2 &= \text{trace} \left[ (X^T \Theta - Y^T)^T \odot \sqrt{W^a} (\sqrt{W^a})^T \odot (X^T \Theta - Y^T) \right] \\
    &= \text{trace} \left[ (\Theta^T X_W - Y_W) (X_W^T \Theta - Y_W^T) \right] \ . 
\end{align*}
We differentiate with respect to $\Theta$ under the $\text{trace}(\cdot)$ operation and obtain
\begin{align*}
    \frac{\partial}{\partial\Theta} |X^T \Theta - Y^T|_{W^a}^2 &= \frac{\partial}{\partial\Theta} \text{trace} \left[ (\Theta^T X_W - Y_W) (X_W^T \Theta - Y_W^T) \right] \\
    &= 2 X_W (X_W ^T\Theta - Y_W^T)  \ ,
\end{align*}
which follows from properties of the $\text{trace}(\cdot)$ and computing the derivative per element. Taking another derivative yields
\[
    \frac{\partial}{\partial\Theta} \left[ 2 X_W (X_W ^T\Theta - Y_W^T) \right] = 2 X_W X_W^T \ ,
\]
which is an SPD matrix as it is the product of a matrix multiplied by the same matrix transposed.

Finally, to guarantee convergence via SGD, we need the gradient $2 X_W (X_W ^T\Theta - Y_W^T)$ to be a Lipschitz function. Indeed, this property holds with a Lipschitz constant of $L = 2 | X_W X_W^T |_F $, where $|\cdot |_F$ is the Frobenius norm. In practice, the data and scaling are bounded and thus $L \ll \infty$.

\section{Variate Grouping}

\subsection{Correlation and Conflict Matrices}

\begin{figure*}[!b]
  \centering
  \includegraphics[width=1.0\linewidth]{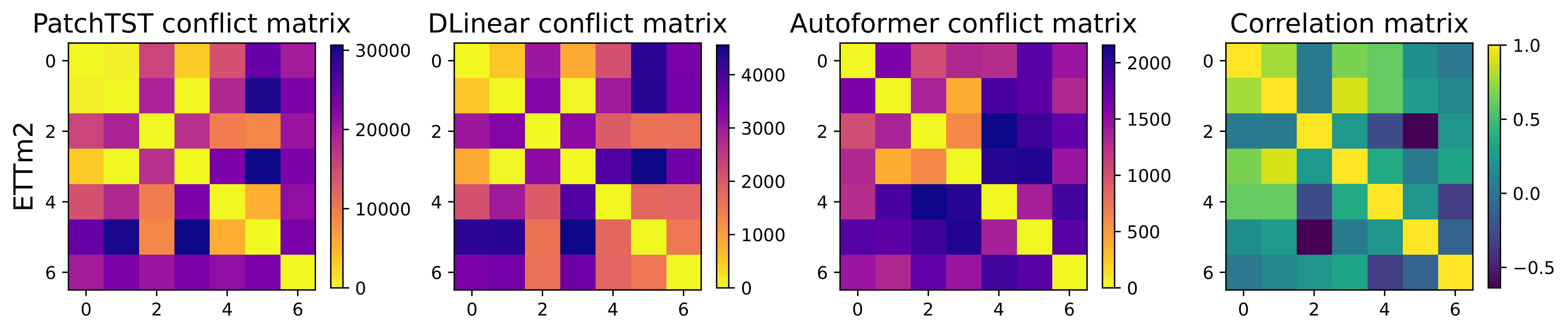}
  \includegraphics[width=1.0\linewidth]{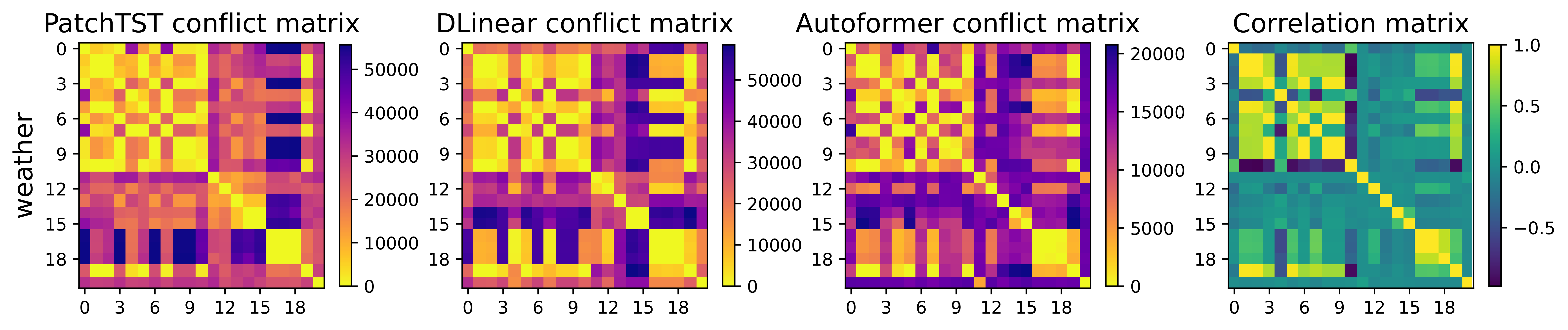}
  \caption{The Pearson correlation matrix (rightmost) and conflict matrices, where each element represents the total number of conflicts seen between two variates during training.}\label{fig:conflict_matrix}
\end{figure*}

In this section, we present the correlation matrices and conflict matrices for the datasets Weather and ETTm2. In Fig.~\ref{fig:conflict_matrix}, each cell in the conflict matrices represents the count of all conflicts between two variates occurred during training. A resemblance is apparent mostly for DLinear and PatchTST. One interesting observation is that strong negative correlations between variates are associated with a low number of conflicts.

\subsection{Task Affinity Grouping (TAG)}
Task affinity grouping (TAG)~\cite{fifty2021efficiently} is a task grouping method, that suggests to select task groups based on the quality of transferability between tasks. In this context, the term \emph{inter-task affinity} encodes the transferability of each task with the remaining tasks during training. TAG proposes two steps. First, obtain a measure of pair-wise task similarity by training a shared model across tasks and observing the contribution of single-task optimization updates on the remaining tasks; the inter-task affinity between tasks is collected and later acts as a measure of task similarity. The next step is calculating the optimal task groups with a selection process that maximizes the
total affinity score. When applied to time series forecasting, certain elements in TAG pose serious difficulties: 1) Training the first steps of this framework could impose a great amount of overhead when applied to the dataset Electricity or Weather since they contain 321 and 21 variates (tasks) respectively. 2) After collecting all the inter-task affinities the selection process maximizes the total affinity score. However, this problem is NP-hard, and thus it requires approximation for using a large number of tasks. Therefore, MTLinear (our approach) can be seen as a more practical approach to variate grouping, and model assignment in TSF settings.


\label{def:hh_clustering}
\subsection{Hierarchical Clustering}
\begin{figure*}[!ht]
  \centering
  \includegraphics[width=1.0\linewidth]{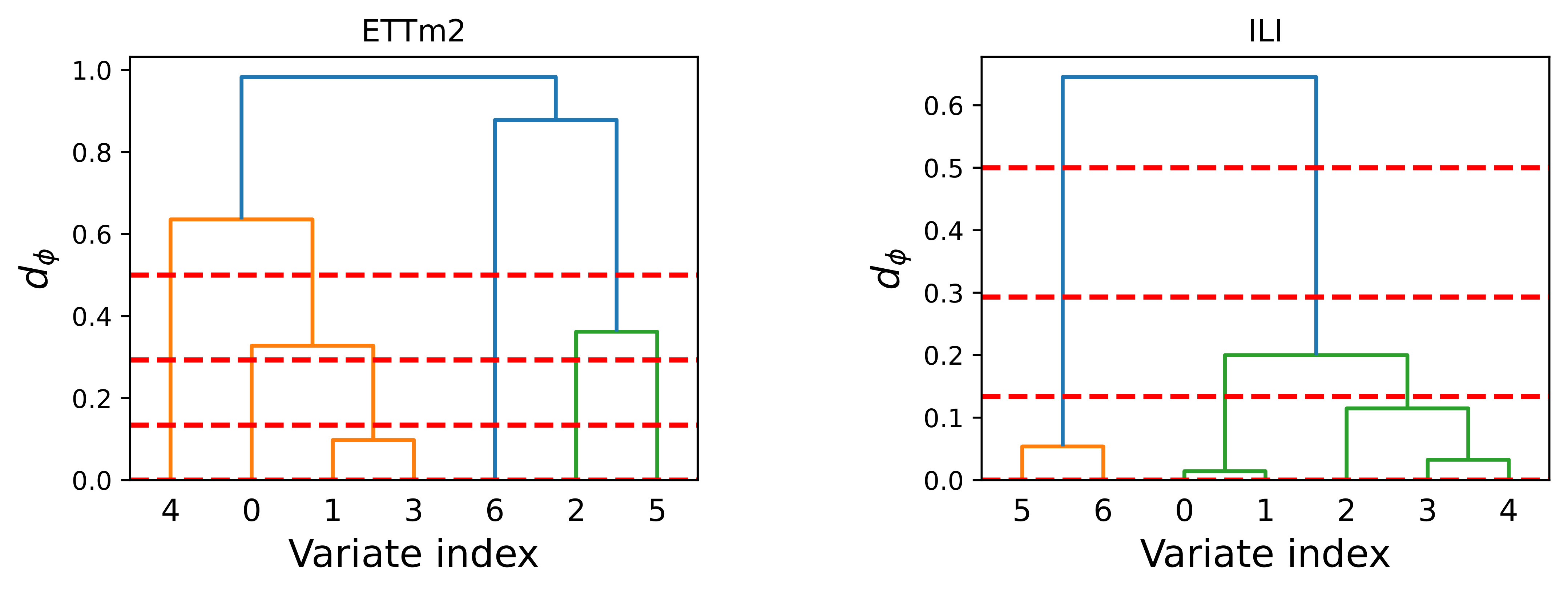}
  \caption{The dendrograms for the ILI and ETTm2 datasets. The red lines represent the cut points $d_{\bar{\alpha}}$ associated with $\bar{\alpha} \in \{\pi/2,\pi/3,\pi/4,\pi/6\}$.}
  \label{fig:dendogram_example}
\end{figure*}

In this section, we briefly present the \emph{hierarchical clustering} algorithm~\cite{hastie2009elements} we utilize in Sec.~\ref{subsec:variate_grouping}. Hierarchical clustering is a general name for a bottom-up (agglomerative) strategy or top-down (divisive) strategy. In this work, we focus on agglomerative clustering which commences with each individual object forming its distinct group. It then systematically combines objects or groups that are proximate to each other until all groups are joined into a single entity at the highest level of the hierarchy or until a termination condition is met. Apart from selecting a metric such as Euclidean distance, a \emph{linkage} criterion must be determined. The linkage defines how to measure the similarity between clusters, which can be defined by the closest pair (\emph{single linkage}), farthest pair (\emph{complete linkage}), or the average of the distances of each observation of the two sets (\emph{average linkage}). A tree structure called a dendrogram is often used to represent the process of hierarchical clustering. We show an example in Fig.~\ref{fig:dendogram_example}.

In our method, we use the Pearson correlation coefficient as the metric between two clusters (variate groups) and employ complete linkage as our linkage strategy. The following table, Table~\ref{tab:number_groups}, presents the number of clusters (groups) for each dataset and criteria $\bar{\alpha}$. We denote the distance between groups by $d_{\bar{\alpha}}$, and we define it as follows, $d_{\bar{\alpha}} = 1 - \cos(\bar{\alpha})$. This term expresses the maximal correlation distance in a given cluster.

\begin{table}[ht]
\caption{Number of variate groups given $\bar{\alpha}$ for each dataset.}
    \centering
    \scalebox{1.0}{
    \begin{tabular}{|l|c|c|c|c|c|}
        \toprule
        Dataset & $\bar{\alpha} = 0$ (max variates) &  $\bar{\alpha} = \pi/6$& $\bar{\alpha} = \pi/4$ & $\bar{\alpha} = \pi/3$ & $\bar{\alpha} = \pi/2$  \\ \midrule
        ECL & $321$ & $175$ & $79$ & $44$ & $1$\\ 
        Traffic & $862$ & $862$ & $838$ & $753$ & $1$\\ 
        Weather & $21$ & $13$ & $11$ & $10$ & $1$\\ 
        ILI & $7$ & $3$ & $2$ & $2$& $1$ \\ 
        ETT & $7$ & $6$ & $6$ & $4$ & $1$\\ 
        Exchange & $8$ & $8$ & $8$ & $7$ & $1$\\
        \bottomrule
    \end{tabular}}
    \label{tab:number_groups}
\end{table}

\subsection{Memory Usage}
\label{app:mem}
Since MTLinear relies on duplicated versions of a single linear model variation, the memory usage of each model can be easily computed according to the number of selected groups. Therefore, Tab.~\ref{tab:number_groups} also describes the memory complexity in units of a single linear model variation memory usage. For example, MTLDinear for Electricity $\bar{\alpha} = \pi/6$ consists of 175 times the memory of a single DLinear model. While this number may seem large, in practice, it is part of a single module implemented in Pytorch where all linear components are stacked such that only the depth of one layer remains. Another benefit of MTLinear is an effective and efficient implementation of quasi-channel independent modeling, where the full memory usage of a channel independent approach ($\bar{\alpha} = 0$) is replaced with a smaller and utilized representation.

\section{Experimental Details}
\label{app:eval_setup}

\paragraph{Datasets.} The proposed method is extensively evaluated on seven common benchmark datasets from different sectors: industrial, weather, energy, and health. The \textbf{Electricity} (ECL) dataset includes the hourly electricity consumption data of 321 customers spanning from 2012 to 2014. \textbf{Weather} is a meteorological dataset recorded every 10 minutes throughout the entire year of 2020, featuring 21 meteorological indicators such as air temperature and humidity. The \textbf{ILI} dataset includes weekly recorded data on influenza-like illness patients from the Centers for Disease Control and Prevention of the United States, spanning from 2002 to 2021. The \textbf{ETT} dataset comprises data gathered from electricity Transformers, encompassing load, and oil temperature readings recorded at 15-minute intervals. \textbf{ETTh2}, \textbf{ETTh1}, \textbf{ETTm2}, and \textbf{ETTm1} form different interval representations, 2 hours, 1 hour, 30 minutes and 15 minutes, respectively.

\paragraph{Baselines.} We selected SOTA and prominent models as the benchmark baselines in our experiments: PatchTST~\cite{nie2023time}, iTransformer~\cite{liu2023itransformer}, Crossformer~\cite{zhang2022crossformer}, FEDformer~\cite{zhou2022fedformer},  Autoformer~\cite{wu2021autoformer}, GPT4TS~\cite{zhou2023one} as well as linear based models Linear, DLinear, NLinear, RLinear~\cite{zeng2023Transformers, li2023revisiting}. The baseline results shown in Tabs.~\ref{tab:main_results_96}, and App.~\ref{tab:main_results_96_full}  are comprised of the reported results in \cite{liu2023itransformer} except for ILI, where the results for Crossformer, FEDformer, and Autoformer were imported from the original paper and the latter was reproduced. In Tab. App.~\ref{tab:main_results_336_full} the results for PatchTST, GPT4TS, DLinear, and NLinear are the original reported results. The remaining experiment tables and figures are reproduced based on the original implementation and hyper-parameters.

\paragraph{Experimental setting.} For all experiments we take the average score of three different seeds. For each seed, we perform grid search and select the setting with the best validation score. The given grid search includes $\bar{\alpha} \in \{ \pi/2,\pi/3,\pi/4,\pi/6\}$ and $a \in \{1,2\}$ for the grouping and penalty parameters, respectively. The other reported results rely on the original implementation and hyperparameters. Most experiments use the standard lookback $l$ of 96, unless mentioned otherwise.

\paragraph{MTLinear implementation details.} MTLinear and its multi-head linear modules are implemented in Pytorch~\cite{paszke2019pytorch}. Similar to the other baselines, we use early stopping. However, we deploy a multi-early stopping scheme, hence, each model's training ends at its own time, thus lifting another form of dependency on other model groups. The maximal number of epochs is set to 20, and learning-rate and batch size are set to $0.01$ and 32, respectively, for all configurations. For the MTLinear setting and other reproduced results, we used RTX 4090 24GB GPU.

\section{Sensitivity Parameter $a$ Ablation}
\label{app:sensitivity_param}
\begin{figure*}[!t]
  \centering
  \includegraphics[width=1.0\linewidth]{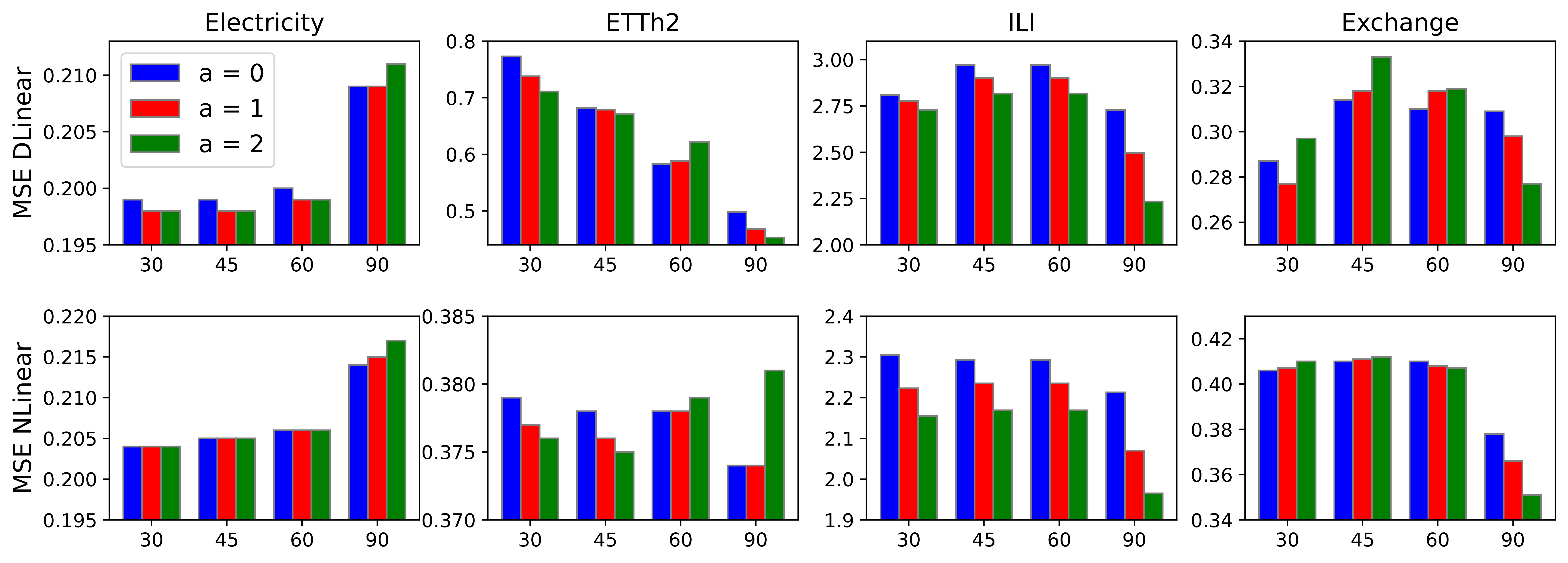}
  \vspace{-5mm}
  \caption{A performance comparison between different values of the sensitivity parameter $a \in \{0,1,2\}$ with respect to each $\bar\alpha\in \{30,45,60,90\}$. }
  \label{fig:sensitivity_param}
\end{figure*}

In this section we provide experiments ablating the effect of $a$ on forecasting, in Fig.~\ref{fig:sensitivity_param}. We consider the values $a$ = [0,1,2]. In most cases, a higher $a$ leads to better results with exceptions. It is also shown that when  $\alpha=90$ the $a$ has a stronger impact on the results, but when variates are grouped that difference is decreased. This behavior can be explained by the fact that very different variates also share large gradient scale differences. In practice, $a$=2 is selected in most cases after hyper-parameter search in the main results.

\section{Extended Main Results}\label{app:extened_results}

This table is the extended version of Tab.~\ref{tab:main_results} with all 4 horizons presented separately. Additionally, we added a table that compares MTLinear and PatchTST \cite{nie2023time} with an input lookback of 336, and GPT4TS \cite{zhou2023one} in Table \ref{tab:main_results_336_full}.

\begin{table*}[ht]
    \caption{Multivariate forecasting results of MTLinear (ours) compared to other strong baselines of different classes. Each score represents one of four forecast horizons $h \in \{24, 36, 48, 60\}$ and 36 input length for ILI, as well as  $h \in \{96, 192, 336, 720\}$ and 96 input length for the remaining. A \textbf{bold} and \underline{underlined} notation represent the best and second-best scores, respectively. }
    \label{tab:main_results}
    \centering
    \resizebox{\linewidth}{!}{
\begin{tabular}{ll|rr|rr|rr|rr|rr|rr|rr|rr|}
\toprule
    &    &  \multicolumn{2}{c|}{MTDLinear} & \multicolumn{2}{c|}{MTNLinear} &  \multicolumn{2}{c|}{iTransformer} &  \multicolumn{2}{c|}{PatchTST} &  \multicolumn{2}{c|}{Crossformer} &  \multicolumn{2}{c|}{DLinear} & \multicolumn{2}{c|}{FEDformer} &  \multicolumn{2}{c|}{Autoformer} \\
\multicolumn{2}{c|}{Dataset}  &       MSE&MAE                  &                 MSE&MAE                    &                      MSE&MAE                         &                 MSE&MAE                     &                     MSE&MAE                        &                 MSE&MAE                    &                   MSE&MAE                      &                    MSE&MAE                       \\
\midrule
\multirow{4}{*}{\rotatebox[origin=c]{90}{ETTm1}} & 96 &            0.337 &            \textbf{0.363} &            0.341 &            0.369 &                \underline{0.334} &                0.368 &            \textbf{0.329} &            \underline{0.367} &               0.404 &               0.426 &           0.345 &           0.372 &             0.379 &             0.419 &              0.505 &              0.475 \\
    & 192 &            0.379 &            0.388 &            0.381 &           \underline{ 0.387} &                \underline{0.377} &                0.391 &            \textbf{0.367} &            \textbf{0.385} &               0.450 &               0.451 &           \underline{0.380} &           0.389 &             0.426 &             0.441 &              0.553 &              0.496 \\
    & 336 &            \underline{0.412} &            0.414 &            0.413 &            \textbf{0.409} &                0.426 &                0.420 &            \textbf{0.399} &            \underline{0.410} &               0.532 &               0.515 &           0.413 &           0.413 &             0.445 &             0.459 &              0.621 &              0.537 \\
    & 720 &            \underline{0.468} &            0.445 &            0.478 &            \underline{0.443} &                0.491 &                0.459 &            \textbf{0.454} &            \textbf{0.439} &               0.666 &               0.589 &           0.474 &           0.453 &             0.543 &             0.490 &              0.671 &              0.561 \\ \cmidrule(lr){1-18}
\multirow{4}{*}{\rotatebox[origin=c]{90}{ETTm2}} & 96 &            \underline{0.179} &             \underline{0.264} &            \textbf{0.175} &            \textbf{0.254} &                0.180 &                 \underline{0.264} &            \textbf{0.175} &            0.259 &               0.287 &               0.366 &           0.193 &           0.292 &             0.203 &             0.287 &              0.255 &              0.339 \\
    & 192 &            0.245 &            0.308 &            \textbf{0.240} &            \textbf{0.296} &                0.250 &                0.309 &           \underline{ 0.241} &            \underline{0.302} &               0.414 &               0.492 &           0.284 &           0.362 &             0.269 &             0.328 &              0.281 &              0.340 \\
    & 336 &            0.306 &            0.350 &            \textbf{0.301} &            \textbf{0.335} &                0.311 &                0.348 &            \underline{0.305} &            \underline{0.343} &               0.597 &               0.542 &           0.369 &           0.427 &             0.325 &             0.366 &              0.339 &              0.372 \\
    & 720 &            \underline{0.407} &            0.415 &            \textbf{0.402} &            \textbf{0.393} &                0.412 &                \underline{0.407} &            \textbf{0.402} &            0.400 &               1.730 &               1.042 &           0.554 &           0.522 &             0.421 &             0.415 &              0.433 &              0.432 \\ \cmidrule(lr){1-18}
\multirow{4}{*}{\rotatebox[origin=c]{90}{ETTh1}} & 96 &            \underline{0.386} &            0.396 &            0.387 &            \textbf{0.393} &                \underline{0.386} &                0.405 &            \underline{0.414} &            0.419 &               0.423 &               0.448 &           \underline{0.386} &           0.400 &             \textbf{0.376} &             0.419 &              0.449 &              0.459 \\
    & 192 &            \underline{0.441} &            \underline{0.426} &            0.439 &            \textbf{0.421} &                0.441 &                0.436 &            0.460 &            0.445 &               0.471 &               0.474 &           0.437 &           0.432 &             \textbf{0.420} &             0.448 &              0.500 &              0.482 \\
    & 336 &            0.490 &            \underline{0.455} &            \underline{0.476} &            \textbf{0.441} &                0.487 &                0.458 &            0.501 &            0.466 &               0.570 &               0.546 &           0.481 &           0.459 &             \textbf{0.459} &             0.465 &              0.521 &              0.496 \\
    & 720 &            0.506 &            \underline{0.488} &            \textbf{0.472} &           \textbf{ 0.460} &                0.503 &                0.491 &            \underline{0.500} &            \underline{0.488} &               0.653 &               0.621 &           0.519 &           0.516 &             0.506 &             0.507 &              0.514 &              0.512 \\ \cmidrule(lr){1-18}
\multirow{4}{*}{\rotatebox[origin=c]{90}{ETTh2}} & 96 &            0.300 &            \underline{0.345} &            \underline{0.288} &            \textbf{0.336} &                \textbf{0.297} &                0.349 &            0.302 &            0.348 &               0.745 &               0.584 &           0.333 &           0.387 &             0.358 &             0.397 &              0.346 &              0.388 \\
    & 192 &            0.390 &            0.405 &            \textbf{0.375} &            \textbf{0.388} &                \underline{0.380} &                \underline{0.400} &            0.388 &            \underline{0.400} &               0.877 &               0.656 &           0.477 &           0.476 &             0.429 &             0.439 &              0.456 &              0.452 \\
    & 336 &            0.500 &            0.482 &            \textbf{0.412} &            \textbf{0.423} &                0.428 &                \underline{0.432} &            \underline{0.42} &            0.433 &               1.043 &               0.731 &           0.594 &           0.541 &             0.496 &             0.487 &              0.482 &              0.486 \\
    & 720 &            0.623 &            0.555 &            \textbf{0.418} &            \textbf{0.440} &                \underline{0.427} &                \underline{0.445} &            0.431 &            0.446 &               1.104 &               0.763 &           0.831 &           0.657 &             0.463 &             0.474 &              0.515 &              0.511 \\ \cmidrule(lr){1-18}
\multirow{4}{*}{\rotatebox[origin=c]{90}{ECL}} & 96 &            0.183 &            0.268 &            0.185 &            \underline{0.265} &                \textbf{0.148} &                \textbf{0.240} &            \underline{0.181} &            0.270 &               0.219 &               0.314 &           0.197 &           0.282 &             0.193 &             0.308 &              0.201 &              0.317 \\
    & 192 &            \underline{0.183} &            0.271 &            0.186 &            \underline{0.268} &                \textbf{0.162} &                \textbf{0.253} &            0.188 &            0.274 &               0.231 &               0.322 &           0.196 &           0.285 &             0.201 &             0.315 &              0.222 &              0.334 \\
    & 336 &            \underline{0.196} &            0.286 &            0.201 &            \underline{0.283} &                \textbf{0.178} &                \textbf{0.269} &            0.204 &            0.293 &               0.246 &               0.337 &           0.209 &           0.301 &             0.214 &             0.329 &              0.231 &              0.338 \\
    & 720 &            \underline{0.231} &            \underline{0.318} &            0.243 &            \textbf{0.317} &                \textbf{0.225} &                \textbf{0.317} &            0.246 &            0.324 &               0.280 &               0.363 &           0.245 &           0.333 &             0.246 &             0.355 &              0.254 &              0.361 \\ \cmidrule(lr){1-18}
\multirow{4}{*}{\rotatebox[origin=c]{90}{Exchange}} & 96 &            \textbf{0.084} &            \underline{0.202} &            \underline{0.085} &            \textbf{0.201} &                0.086 &                0.206 &            0.088 &            0.205 &               0.256 &               0.367 &           0.088 &           0.218 &             0.148 &             0.278 &              0.197 &              0.323 \\
    & 192 &            \textbf{0.173} &            \underline{0.300} &            0.183 &            0.302 &                0.177 &                \textbf{0.299} &            \underline{0.176} &            \textbf{0.299} &               0.470 &               0.509 &           0.176 &           0.315 &             0.271 &             0.315 &              0.300 &              0.369 \\
    & 336 &           \underline{0.306} &            \underline{0.412} &            0.355 &            0.428 &                0.331 &                0.417 &            \textbf{0.301} &            \textbf{0.397} &               1.268 &               0.883 &           0.313 &           0.427 &             0.460 &             0.427 &              0.509 &              0.524 \\
    & 720 &           \textbf{0.595} &            \textbf{0.595} &            1.015 &            0.758 &                0.847 &                \underline{0.691} &            0.901 &            0.714 &               1.767 &               1.068 &           \underline{0.839} &           0.695 &             1.195 &             0.695 &              1.447 &              0.941 \\ \cmidrule(lr){1-18}
\multirow{4}{*}{\rotatebox[origin=c]{90}{Traffic}} & 96 &            0.648 &            \underline{0.396} &            0.647 &            0.383 &                \textbf{0.395} &                \textbf{0.268} &            \underline{0.462} &            \underline{0.295} &               0.522 &               0.290 &           0.650 &           0.396 &             0.587 &             0.366 &              0.613 &              0.388 \\
    & 192 &            0.594 &            0.365 &            0.598 &            0.359 &                \textbf{0.417} &                \textbf{0.276} &            \underline{0.466} &            \underline{0.296} &               0.530 &               0.293 &           0.598 &           0.370 &             0.604 &             0.373 &              0.616 &              0.382 \\
    & 336 &            0.601 &            0.368 &            0.606 &            0.362 &                \textbf{0.433} &                \textbf{0.283} &            \underline{0.482} &            \underline{0.304} &               0.558 &               0.305 &           0.605 &           0.373 &             0.621 &             0.383 &              0.622 &              0.337 \\
    & 720 &            0.640 &            0.393 &            0.644 &            0.382 &               \textbf{0.467} &                \textbf{0.302} &           \underline{ 0.514} &            \underline{0.322} &               0.589 &               0.328 &           0.645 &           0.394 &             0.626 &             0.382 &              0.660 &              0.408 \\ \cmidrule(lr){1-18}
\multirow{4}{*}{\rotatebox[origin=c]{90}{Weather}} & 96 &            \underline{0.159} &            0.221 &            0.166 &            \textbf{0.211} &                0.174 &                \underline{0.214} &            0.177 &            0.218 &               \textbf{0.158} &               0.230 &           0.196 &           0.255 &             0.217 &             0.296 &              0.266 &              0.336 \\
    & 192 &           \textbf{ 0.202} &            0.268 &            0.212 &            \textbf{0.252} &                0.221 &                \underline{0.254} &            0.225 &            0.259 &               \underline{0.206} &               0.277 &           0.237 &           0.296 &             0.276 &             0.336 &              0.307 &              0.367 \\
    & 336 &            \textbf{0.259} &            0.318 &            \underline{0.268} &            \textbf{0.294} &                0.278 &               \underline{ 0.296} &            0.278 &            0.297 &               0.272 &               0.335 &           0.283 &           0.335 &             0.339 &             0.380 &              0.359 &              0.395 \\
    & 720 &            \textbf{0.332} &            0.373 &            0.349 &            \textbf{0.346} &                0.358 &                \underline{0.347} &            0.354 &            0.348 &               0.398 &               0.418 &           \underline{0.345} &           0.381 &             0.403 &             0.428 &              0.419 &              0.428 \\ \cmidrule(lr){1-18}
\multirow{4}{*}{\rotatebox[origin=c]{90}{ILI}} & 24 &            \underline{2.246} &            \underline{0.993} &            \textbf{2.126} &            \textbf{0.928} &                2.754 &                1.103 &            2.390 &            0.999 &               3.041 &               1.186 &           2.398 &           1.040 &             3.228 &             1.260 &              3.483 &              1.287 \\
    & 36 &            \underline{2.233} &            \underline{0.989} &            \textbf{1.914} &            \textbf{0.886} &                2.707 &                1.074 &            2.331 &            0.994 &               3.406 &               1.232 &           2.646 &           1.088 &             2.679 &             1.080 &              3.103 &              1.148 \\
    & 48 &            \underline{2.102} &            \underline{0.962} &            \textbf{1.795} &            \textbf{0.867} &                2.610 &                1.068 &            2.488 &            1.033 &               3.459 &               1.221 &           2.614 &           1.086 &             2.622 &             1.078 &              2.669 &              1.085 \\
    & 60 &            \underline{2.357} &            1.036 &            \textbf{2.023} &           \textbf{0.927} &                2.881 &                1.147 &            2.475 &            \underline{1.018} &               3.640 &               1.305 &           2.804 &           1.146 &             2.857 &             1.157 &              2.770 &              1.125 \\
\cmidrule(lr){1-18}
 \multicolumn{2}{c|}{Average}& \underline{0.575} &            0.440 &            \textbf{0.550} &            \textbf{0.422} &                0.611 &                0.436 &            0.584 &            \underline{0.431} &               0.902 &               0.579 &           0.649 &           0.475 &             0.681 &             0.483 &              0.741 &              0.515\\
 
 \multicolumn{2}{c|}{$1^{st}$ Count} &            6 &            2 &            \textbf{12} &            \textbf{23} &                \underline{9} &                \underline{9} &            7 &            4 &               1 &               0 &           0 &           0 &             3 &             0 &              0 &              0 \\
\bottomrule
\label{tab:main_results_96_full}
\end{tabular}}
\end{table*}

\begin{table}[!b]
    \caption{Multivariate forecasting results of MTLinear (ours) compared to other strong baselines of different classes. Each score represents one of four forecast horizons $h \in \{24, 36, 48, 60\}$ and 104 input length for ILI, as well as  $h \in \{96, 192, 336, 720\}$ and 336 input length for the remaining. A \textbf{bold} and \underline{underlined} notation represent the best and second-best scores, respectively. }
    \centering
    \resizebox{.8\linewidth}{!}{
\begin{tabular}{lc|c|c|c|c|c|c}
\toprule
      & Class& \multicolumn{2}{|c|}{MTLinear } & \multicolumn{2}{c|}{Transformer} & \multicolumn{2}{c}{Linear}  \\
     \midrule
Dataset &  &      MTDLinear        &    MTNLinear   &    GPT4TS    & PatchTST&   DLinear & NLinear\\
\midrule
\multirow{4}{*}{\rotatebox[origin=c]{90}{ETTm2}} & 96  &        0.166 &        \textbf{0.160} &  0.173 &     \underline{0.165} &          0.167 &    0.167 \\
      & 192 &        0.222 &        \textbf{0.216} &  0.229 &     \underline{0.220} &           0.224 &    0.221 \\
      & 336 &        0.285 &       \textbf{0.272} &  0.286 &     0.278 &          0.281 &    \underline{0.274} \\
      & 720 &        0.377 &        \textbf{0.367} &  0.378 &    \textbf{ 0.367} &           0.397 &    \underline{0.368} \\ \hline
\multirow{4}{*}{\rotatebox[origin=c]{90}{ECL}} & 96  &        \underline{0.133} &        0.134 &  0.139 &     \textbf{0.130} &           0.140 &    0.141 \\
      & 192 &        \textbf{0.148} &       \underline{ 0.149} &  0.153 &     \textbf{0.148} &           0.153 &    0.154 \\
      & 336 &        \textbf{0.164} &        \underline{0.167} &  0.169 &     \underline{0.167} &           0.169 &    0.171 \\
      & 720 &        \textbf{0.199} &        0.205 &  0.206 &     \underline{0.202} &         0.203 &    0.210 \\ \hline
\multirow{4}{*}{\rotatebox[origin=c]{90}{Weather}} & 96  &        \textbf{0.145} &        \underline{0.146} &  0.162 &     0.152 &        0.176 &    0.182 \\
      & 192 &        \textbf{0.187} &        \underline{0.189} &  0.204 &     0.197 &           0.220 &    0.225 \\
      & 336 &        \textbf{0.238} &        \underline{0.241} &  0.254 &     0.249 &           0.265 &    0.271 \\
      & 720 &        \textbf{0.310} &        \underline{0.320} &  0.326 &     \underline{0.320} &           0.323 &    0.338 \\ \hline
\multirow{4}{*}{\rotatebox[origin=c]{90}{ILI}} & 24  &        2.024 &        \underline{1.611} &  2.063 &    \textbf{1.522}  &           2.215 &    1.683 \\
      & 36  &        2.063 &        \underline{1.546} &  1.868 &     \textbf{1.430} &          1.963 &    1.703 \\
      & 48  &        2.086 &        \textbf{1.554} &  1.790 &     \underline{1.673} &          2.130 &    1.719 \\
      & 60  &        2.256 &        \underline{1.735} &  1.979 &     \textbf{1.529} &           2.368 &    1.819 \\ \hline
      \multicolumn{2}{c|}{$1^{st}$ Count}  &       \textbf{ 7} &        5 &  0 &     \underline{6}  &    0 &    0 \\

\bottomrule
\label{tab:main_results_336_full}
\end{tabular}}
\end{table}

\clearpage

\subsection{Main results with standard deviation}
 In Tab.~\ref{tab:main_results_96_std} the main results for MTLinear are presented alongside the standard deviation. We should note that the standard deviation for other baseline models is not given here since the corresponding results were taken from other papers, mainly \cite{liu2023itransformer}.

\begin{table}[ht]
    \caption{Multivariate forecasting results of MTLinear (ours) with the standard deviation, Each score represents one of four forecast horizons $h \in \{24, 36, 48, 60\}$ and 36 input length for ILI, as well as  $h \in \{96, 192, 336, 720\}$ and 96 input length for the remaining.}
    \centering
    \resizebox{.65\linewidth}{!}{
\begin{tabular}{ll|ll|ll|}
\toprule
    &    &  \multicolumn{2}{c|}{MTDLinear} & \multicolumn{2}{c|}{MTNLinear} \\
\multicolumn{2}{c|}{Dataset}  &       MSE&MAE                  &                 MSE&MAE \\                  
\midrule
\multirow{4}{*}{\rotatebox[origin=c]{90}{ETTm1}} & 96 &   0.337 ± 0.001 &   0.363 ± 0.002 &   0.341 ± 0.002 &   0.369 ± 0.001 \\
    & 192 &   0.379 ± 0.002 &   0.388 ± 0.005 &   0.381 ± 0.001 &   0.387 ± 0.001 \\
    & 336 &   0.412 ± 0.003 &   0.414 ± 0.006 &   0.413 ± 0.001 &   0.409 ± 0.001 \\
    & 720 &   0.468 ± 0.002 &   0.445 ± 0.001 &   0.478 ± 0.001 &   0.443 ± 0.001 \\ \hline
\multirow{4}{*}{\rotatebox[origin=c]{90}{ETTm2}} & 96 &     0.179 ± 0.0 &   0.264 ± 0.001 &     0.175 ± 0.0 &     0.254 ± 0.0 \\
    & 192 &     0.245 ± 0.0 &   0.308 ± 0.001 &      0.24 ± 0.0 &     0.296 ± 0.0 \\
    & 336 &     0.306 ± 0.0 &    0.35 ± 0.002 &     0.301 ± 0.0 &     0.335 ± 0.0 \\
    & 720 &   0.407 ± 0.005 &    0.415 ± 0.01 &     0.402 ± 0.0 &     0.393 ± 0.0 \\ \hline
\multirow{4}{*}{\rotatebox[origin=c]{90}{ETTh1}} & 96 &   0.386 ± 0.003 &   0.396 ± 0.004 &   0.387 ± 0.002 &   0.393 ± 0.002 \\
    & 192 &   0.441 ± 0.005 &   0.426 ± 0.005 &     0.439 ± 0.0 &     0.421 ± 0.0 \\
    & 336 &    0.49 ± 0.008 &   0.455 ± 0.008 &   0.476 ± 0.001 &     0.441 ± 0.0 \\
    & 720 &   0.506 ± 0.007 &   0.488 ± 0.006 &   0.472 ± 0.001 &      0.46 ± 0.0 \\ \hline
\multirow{4}{*}{\rotatebox[origin=c]{90}{ETTh2}} & 96 &     0.3 ± 0.001 &   0.345 ± 0.004 &   0.288 ± 0.001 &     0.336 ± 0.0 \\
    & 192 &    0.39 ± 0.004 &   0.405 ± 0.004 &     0.375 ± 0.0 &     0.388 ± 0.0 \\
    & 336 &     0.5 ± 0.017 &   0.482 ± 0.011 &   0.412 ± 0.001 &     0.423 ± 0.0 \\
    & 720 &   0.623 ± 0.044 &    0.555 ± 0.02 &   0.418 ± 0.002 &    0.44 ± 0.001 \\ \hline
\multirow{4}{*}{\rotatebox[origin=c]{90}{ECL}} & 96 &     0.183 ± 0.0 &     0.268 ± 0.0 &     0.185 ± 0.0 &     0.265 ± 0.0 \\
    & 192 &     0.183 ± 0.0 &     0.271 ± 0.0 &     0.186 ± 0.0 &     0.268 ± 0.0 \\
    & 336 &     0.196 ± 0.0 &     0.286 ± 0.0 &     0.201 ± 0.0 &     0.283 ± 0.0 \\
    & 720 &     0.231 ± 0.0 &   0.318 ± 0.001 &     0.243 ± 0.0 &     0.317 ± 0.0 \\ \hline
\multirow{4}{*}{\rotatebox[origin=c]{90}{Exchange}} & 96 &   0.084 ± 0.003 &   0.202 ± 0.003 &   0.085 ± 0.005 &   0.201 ± 0.005 \\
    & 192 &    0.173 ± 0.01 &     0.3 ± 0.008 &   0.183 ± 0.001 &   0.302 ± 0.002 \\
    & 336 &   0.306 ± 0.003 &   0.412 ± 0.003 &   0.355 ± 0.003 &   0.428 ± 0.002 \\
    & 720 &   0.595 ± 0.156 &    0.595 ± 0.06 &   1.015 ± 0.009 &   0.758 ± 0.004 \\ \hline
\multirow{4}{*}{\rotatebox[origin=c]{90}{Traffic}} & 96 &     0.648 ± 0.0 &     0.396 ± 0.0 &     0.647 ± 0.0 &     0.383 ± 0.0 \\
    & 192 &     0.594 ± 0.0 &     0.365 ± 0.0 &     0.598 ± 0.0 &     0.359 ± 0.0 \\
    & 336 &     0.601 ± 0.0 &     0.368 ± 0.0 &     0.606 ± 0.0 &     0.362 ± 0.0 \\
    & 720 &      0.64 ± 0.0 &     0.393 ± 0.0 &     0.644 ± 0.0 &     0.382 ± 0.0 \\ \hline
\multirow{4}{*}{\rotatebox[origin=c]{90}{Weather}} & 96 &   0.159 ± 0.001 &   0.221 ± 0.001 &   0.166 ± 0.002 &   0.211 ± 0.001 \\
    & 192 &     0.202 ± 0.0 &   0.268 ± 0.001 &     0.212 ± 0.0 &     0.252 ± 0.0 \\
    & 336 &   0.259 ± 0.005 &   0.318 ± 0.006 &   0.268 ± 0.001 &     0.294 ± 0.0 \\
    & 720 &   0.332 ± 0.005 &   0.373 ± 0.007 &     0.349 ± 0.0 &     0.346 ± 0.0 \\ \hline
\multirow{4}{*}{\rotatebox[origin=c]{90}{ILI}} & 24 &   2.246 ± 0.043 &   0.993 ± 0.017 &   2.126 ± 0.027 &   0.928 ± 0.007 \\ 
    & 36 &   2.233 ± 0.123 &   0.989 ± 0.049 &   1.914 ± 0.074 &   0.886 ± 0.028 \\
    & 48 &   2.102 ± 0.058 &    0.962 ± 0.01 &   1.795 ± 0.008 &   0.867 ± 0.005 \\
    & 60 &   2.357 ± 0.021 &   1.036 ± 0.008 &   2.023 ± 0.007 &   0.927 ± 0.003 \\ 
\bottomrule
\label{tab:main_results_96_std}
\end{tabular}}
\end{table}

\subsection{Gradient manipulation full table results}
In this subsection, the full result tables for DLinear \ref{tab:full_grad_manip_DLinear} and NLinear \ref{tab:full_grad_manip_NLinear}, these results extend Tab.~\ref{tab:grad_manip} in the main text. Each score represents the average result of three different runs corresponding to different seeds which offer the best validation score after grid search for different parameters.

\begin{table}[t]
    \caption{DLinear: Multivariate forecasting results of Gradient manipulation methods, Each score represents one of four forecast horizons $h \in \{24, 36, 48, 60\}$ and 36 input length for ILI, as well as  $h \in \{96, 192, 336, 720\}$ and 96 input length for the remaining.}
    \centering
    \resizebox{.75\linewidth}{!}{
\begin{tabular}{ll|c|c|c|c|c|}
\toprule
   DLinear     &       &  GradNorm &  Cov-W &  PCgrad &  CAGrad &  Nash-MTL \\
Dataset & Horizon &                       &                  &                     &                     &                      \\
\midrule
 \hline 
 \multirow{4}{*}{\rotatebox[origin=c]{90}{ETTm1}} & 96  &                 0.384 &            0.344 &               0.348 &               0.345 &                0.344 \\
        & 192 &                 0.463 &            0.381 &               0.384 &               0.382 &                0.382 \\
        & 336 &                 0.468 &            0.416 &               0.417 &               0.418 &                0.418 \\
        & 720 &                 0.573 &            0.482 &               0.494 &               0.480 &                0.480 \\
 \hline 
 \multirow{4}{*}{\rotatebox[origin=c]{90}{ETTm2}} & 96  &                 0.261 &            0.184 &               0.181 &               0.183 &                0.188 \\
        & 192 &                 0.374 &            0.263 &               0.249 &               0.255 &                0.273 \\
        & 336 &                 0.458 &            0.337 &               0.321 &               0.335 &                0.363 \\
        & 720 &                 0.694 &            0.482 &               0.458 &               0.451 &                0.477 \\
 \hline 
 \multirow{4}{*}{\rotatebox[origin=c]{90}{Exchange}} & 96  &                 0.122 &            0.080 &               0.087 &               0.120 &                0.120 \\
        & 192 &                 0.229 &            0.164 &               0.157 &               0.207 &                0.207 \\
        & 336 &                 0.508 &            0.289 &               0.316 &               0.340 &                0.340 \\
        & 720 &                 0.656 &            0.712 &               0.740 &               0.833 &                0.833 \\
 \hline 
 \multirow{4}{*}{\rotatebox[origin=c]{90}{ILI}} & 24  &                 4.765 &            2.576 &               2.414 &               2.365 &                2.710 \\
        & 36  &                 6.774 &            2.517 &               2.435 &               2.336 &                2.835 \\
        & 48  &                 2.998 &            2.528 &               2.347 &               2.301 &                2.682 \\
        & 60  &                 4.862 &            2.777 &               2.576 &               2.396 &                2.895 \\
 \hline 
 \multirow{4}{*}{\rotatebox[origin=c]{90}{Weather}} & 96  &                 0.203 &            0.196 &               0.196 &               0.196 &                0.197 \\
        & 192 &                 0.253 &            0.237 &               0.236 &               0.238 &                0.238 \\
        & 336 &                 0.301 &            0.284 &               0.286 &               0.285 &                0.285 \\
        & 720 &                 0.361 &            0.354 &               0.348 &               0.348 &                0.348 \\
\bottomrule
\end{tabular}}
\label{tab:full_grad_manip_DLinear}
\end{table}

\begin{table}[t]
    \caption{NLinear: Multivariate forecasting results of Gradient manipulation methods, Each score represents one of four forecast horizons $h \in \{24, 36, 48, 60\}$ and 36 input length for ILI, as well as  $h \in \{96, 192, 336, 720\}$ and 96 input length for the remaining.}
    \centering
    \resizebox{.75\linewidth}{!}{
\begin{tabular}{ll|c|c|c|c|c|}
\toprule
 NLinear       &       &  GradNorm &  Cov-W &  PCgrad &  CAGrad &  Nash-MTL \\
Dataset & Horizon &                       &                  &                     &                     &                      \\
\midrule
 \hline 
 \multirow{4}{*}{\rotatebox[origin=c]{90}{ETTm1}} & 96  &                 0.408 &            0.349 &               0.354 &               0.351 &                0.355 \\
        & 192 &                 0.409 &            0.388 &               0.393 &               0.396 &                0.389 \\
        & 336 &                 0.437 &            0.421 &               0.424 &               0.424 &                0.423 \\
        & 720 &                 0.499 &            0.483 &               0.484 &               0.486 &                0.486 \\
 \hline 
 \multirow{4}{*}{\rotatebox[origin=c]{90}{ETTm2}} & 96  &                 0.203 &            0.182 &               0.181 &               0.180 &                0.183 \\
        & 192 &                 0.256 &            0.245 &               0.245 &               0.245 &                0.245 \\
        & 336 &                 0.318 &            0.306 &               0.306 &               0.305 &                0.306 \\
        & 720 &                 0.417 &            0.407 &               0.406 &               0.405 &                0.407 \\
 \hline 
 \multirow{4}{*}{\rotatebox[origin=c]{90}{Exchange}} & 96  &                 0.088 &            0.088 &               0.088 &               0.085 &                0.085 \\
        & 192 &                 0.190 &            0.178 &               0.170 &               0.175 &                0.175 \\
        & 336 &                 0.356 &            0.330 &               0.322 &               0.319 &                0.319 \\
        & 720 &                 1.044 &            0.922 &               0.904 &               0.851 &                0.851 \\
 \hline 
 \multirow{4}{*}{\rotatebox[origin=c]{90}{ILI}} & 24  &                 2.933 &            2.351 &               2.327 &               2.191 &                2.348 \\
        & 36  &                 2.257 &            2.088 &               2.103 &               1.907 &                2.136 \\
        & 48  &                 2.426 &            2.040 &               1.999 &               1.823 &                2.072 \\
        & 60  &                 2.546 &            2.216 &               2.170 &               2.028 &                2.249 \\
 \hline 
 \multirow{4}{*}{\rotatebox[origin=c]{90}{Weather}} & 96  &                 0.247 &            0.195 &               0.193 &               0.195 &                0.195 \\
        & 192 &                 0.279 &            0.241 &               0.239 &               0.241 &                0.241 \\
        & 336 &                 0.311 &            0.293 &               0.292 &               0.293 &                0.293 \\
        & 720 &                 0.388 &            0.365 &               0.364 &               0.365 &                0.366 \\
\bottomrule
\end{tabular}}
\label{tab:full_grad_manip_NLinear}
\end{table}

\clearpage

\subsection{PCA of the time series datasets}
\begin{figure}[t]
    \centering
    \includegraphics[width=0.7\textwidth]{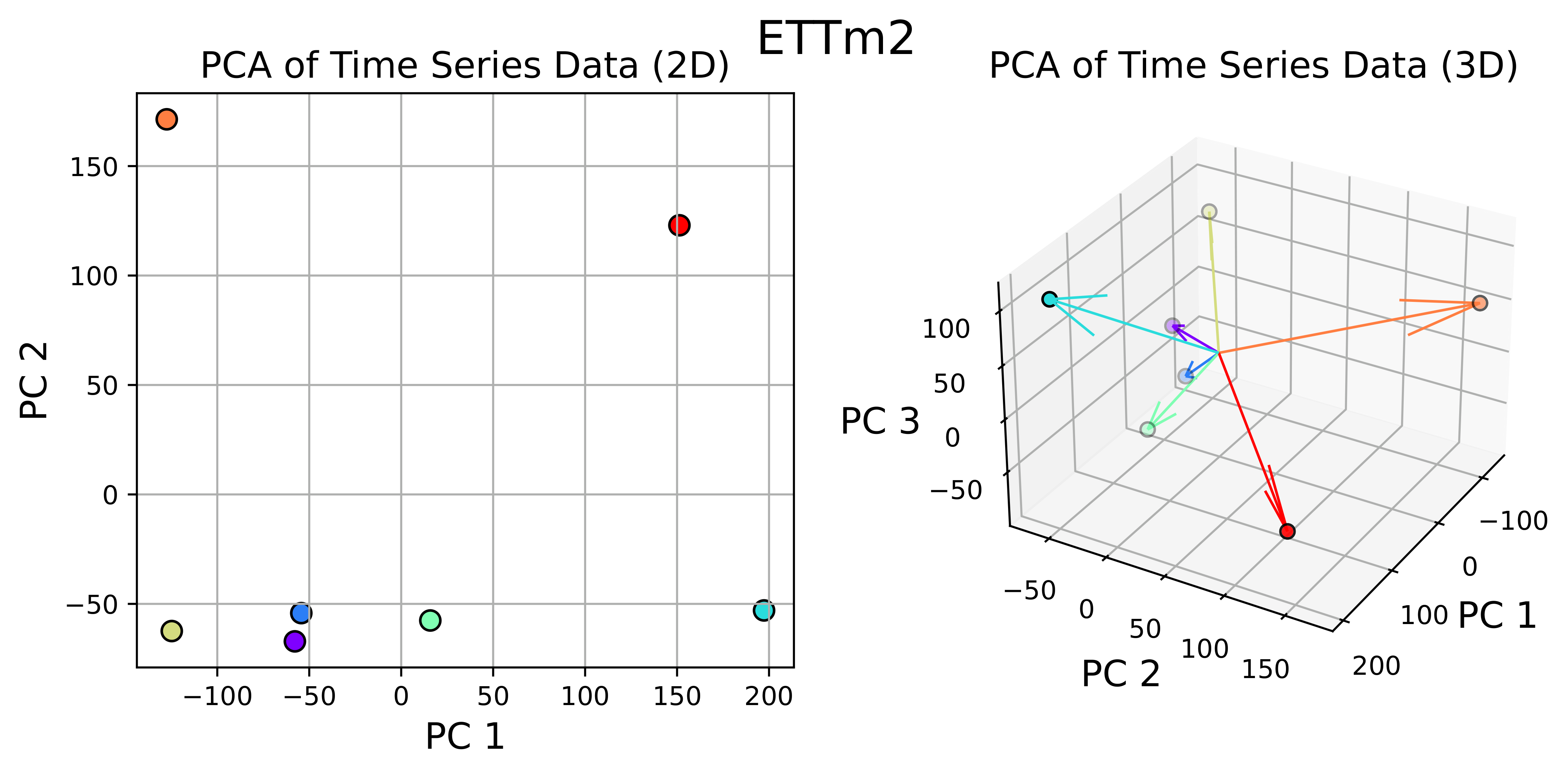} 
    \includegraphics[width=0.7\textwidth]{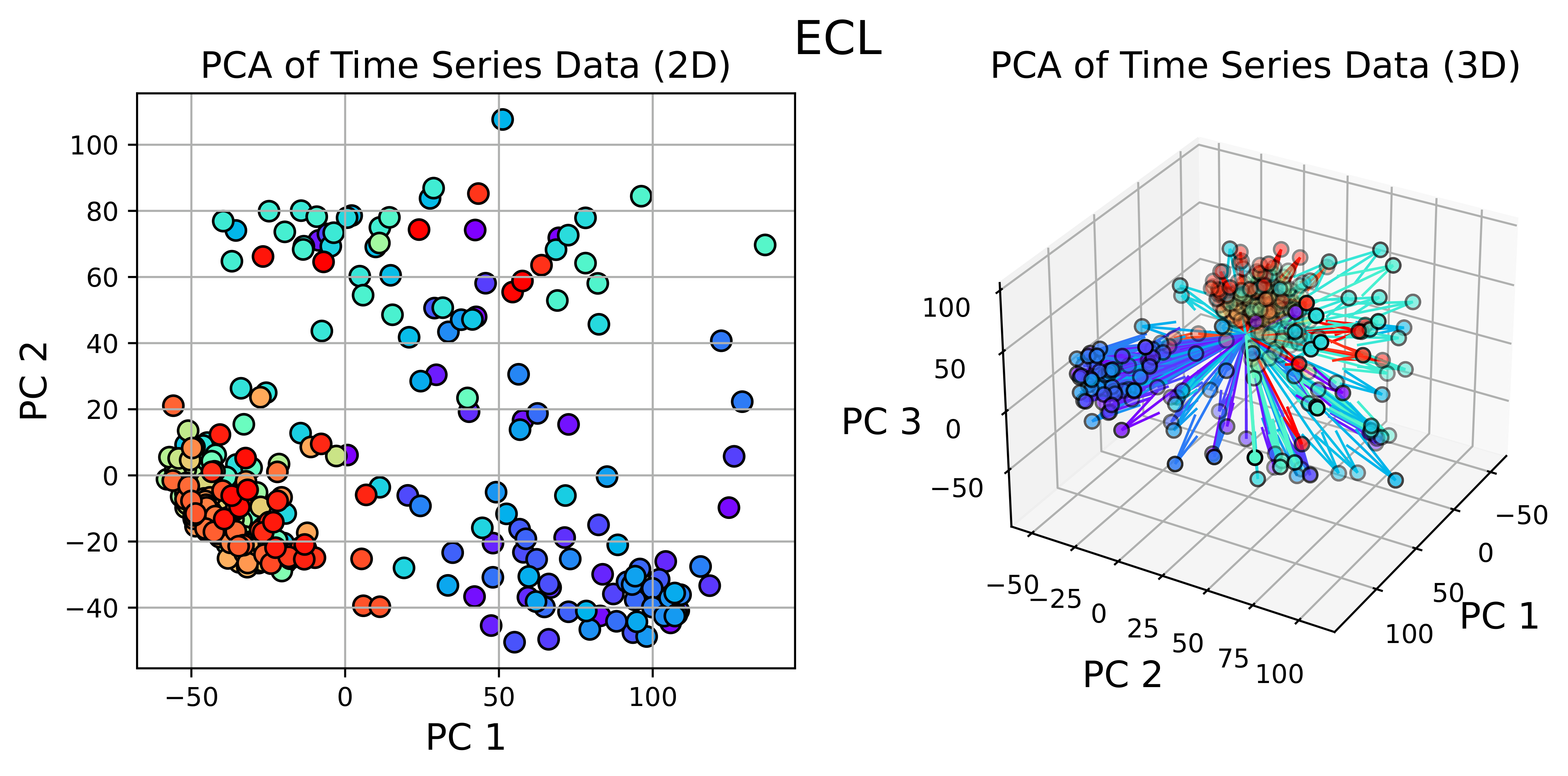} 
    \includegraphics[width=0.7\textwidth]{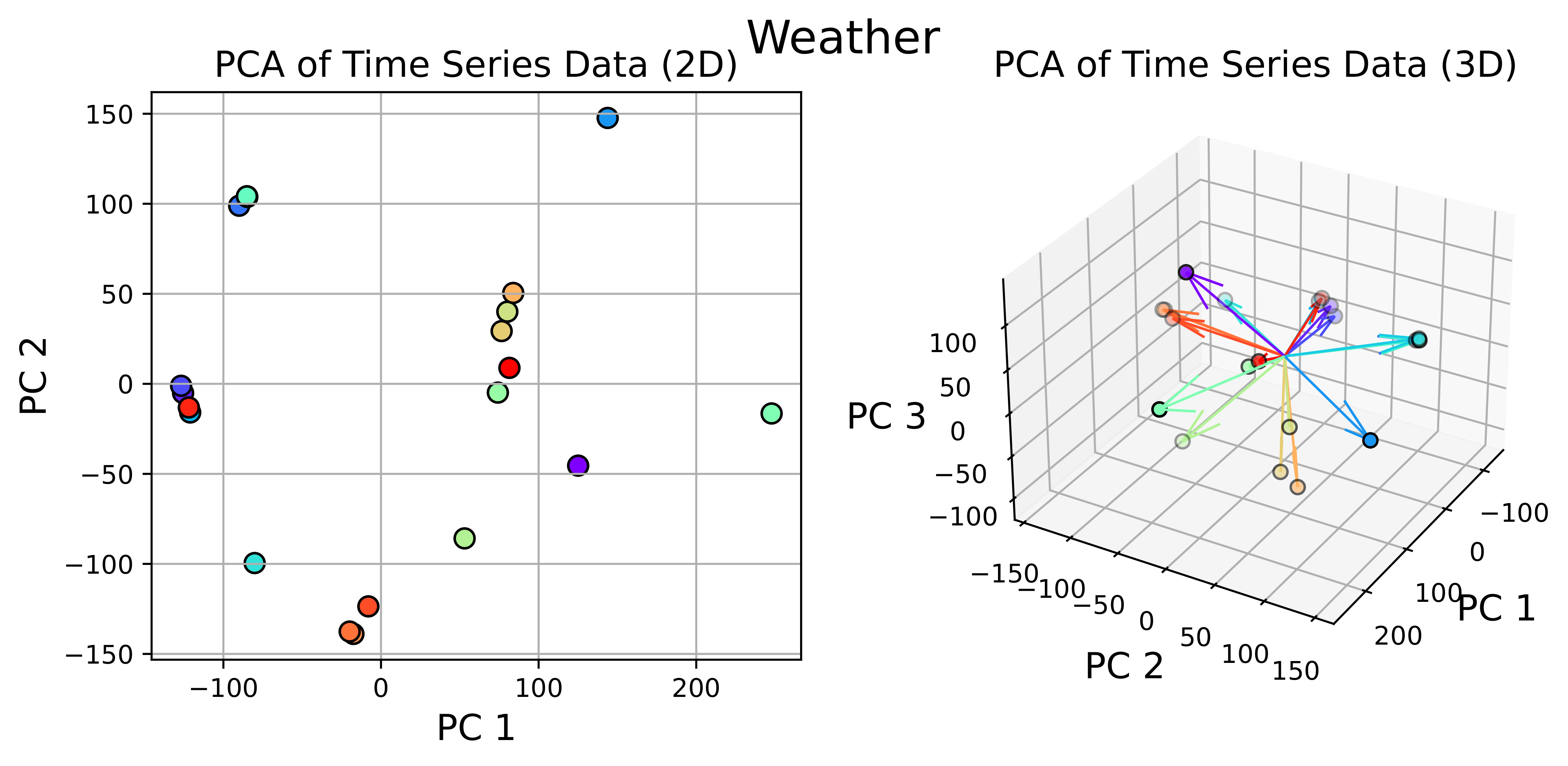} 
    \caption{2D (left) and 3D (right) PCA applied to ETTm2, ECL, and Weather datasets. PCA assists with highlighting variate direction and similarity.}
    \label{fig:PCA_1}
\end{figure}

\begin{figure}[t]
    \centering
    \includegraphics[width=0.7\textwidth]{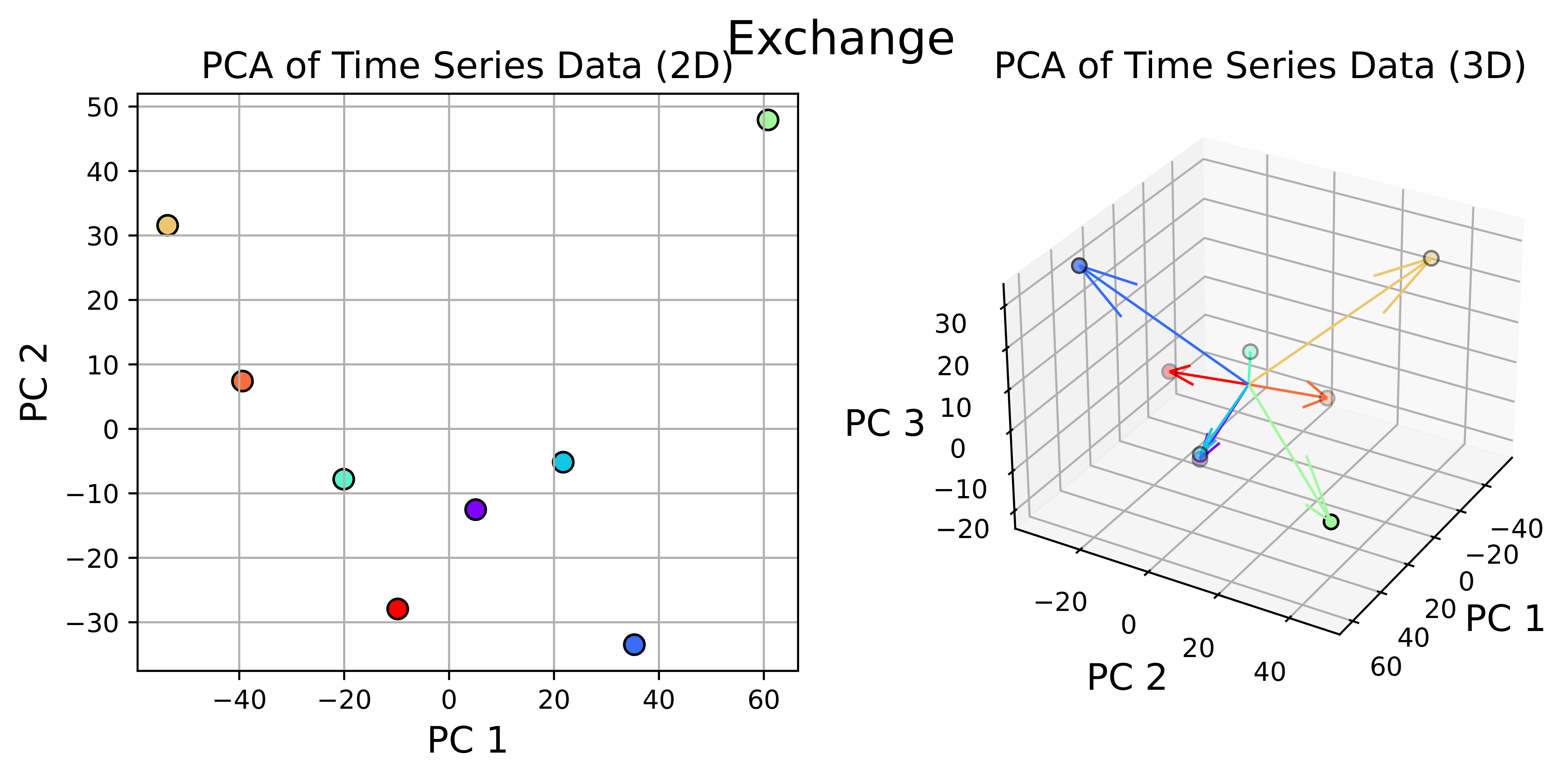} 
    \includegraphics[width=0.7\textwidth]{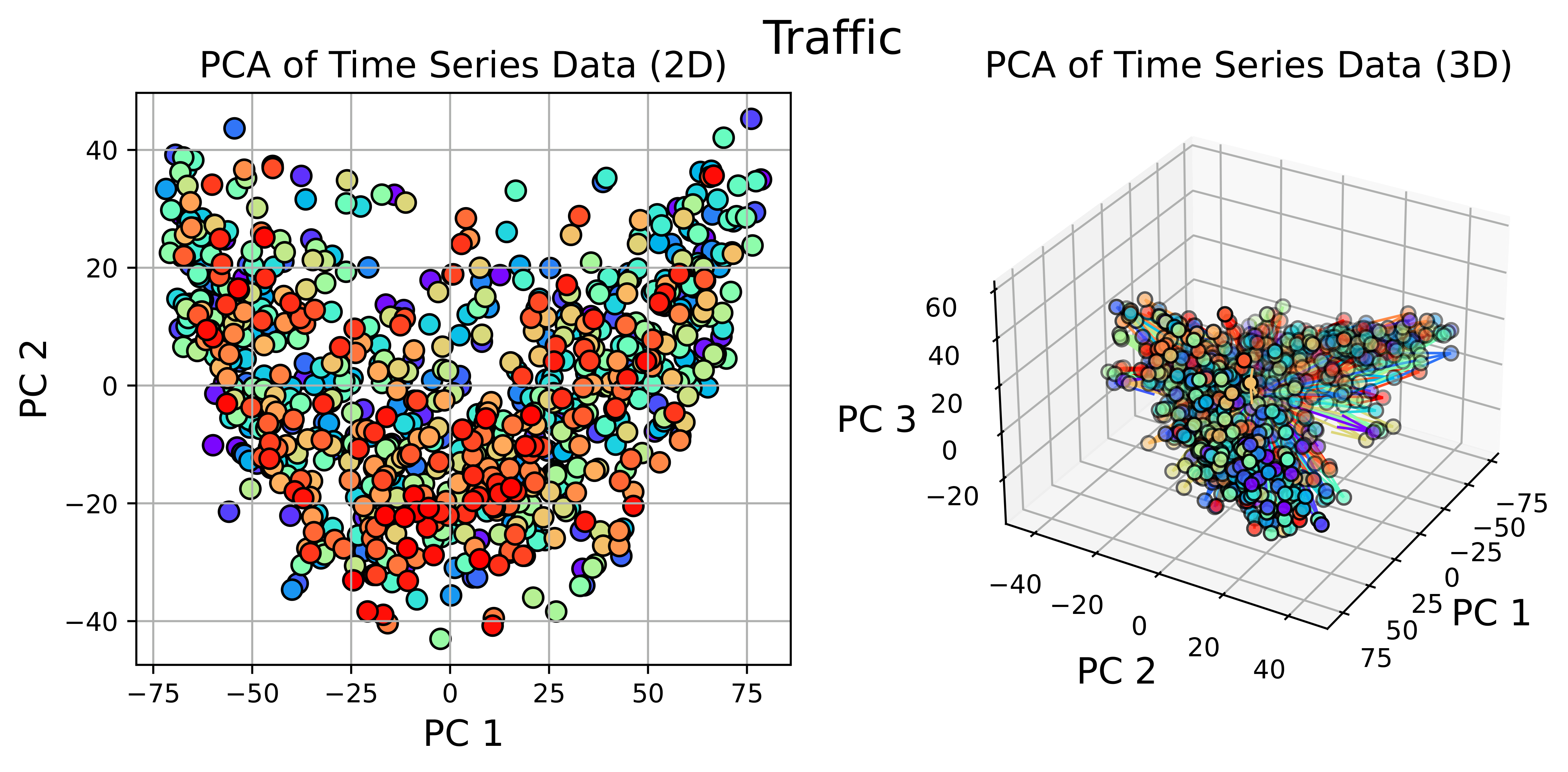} 
    \includegraphics[width=0.7\textwidth]{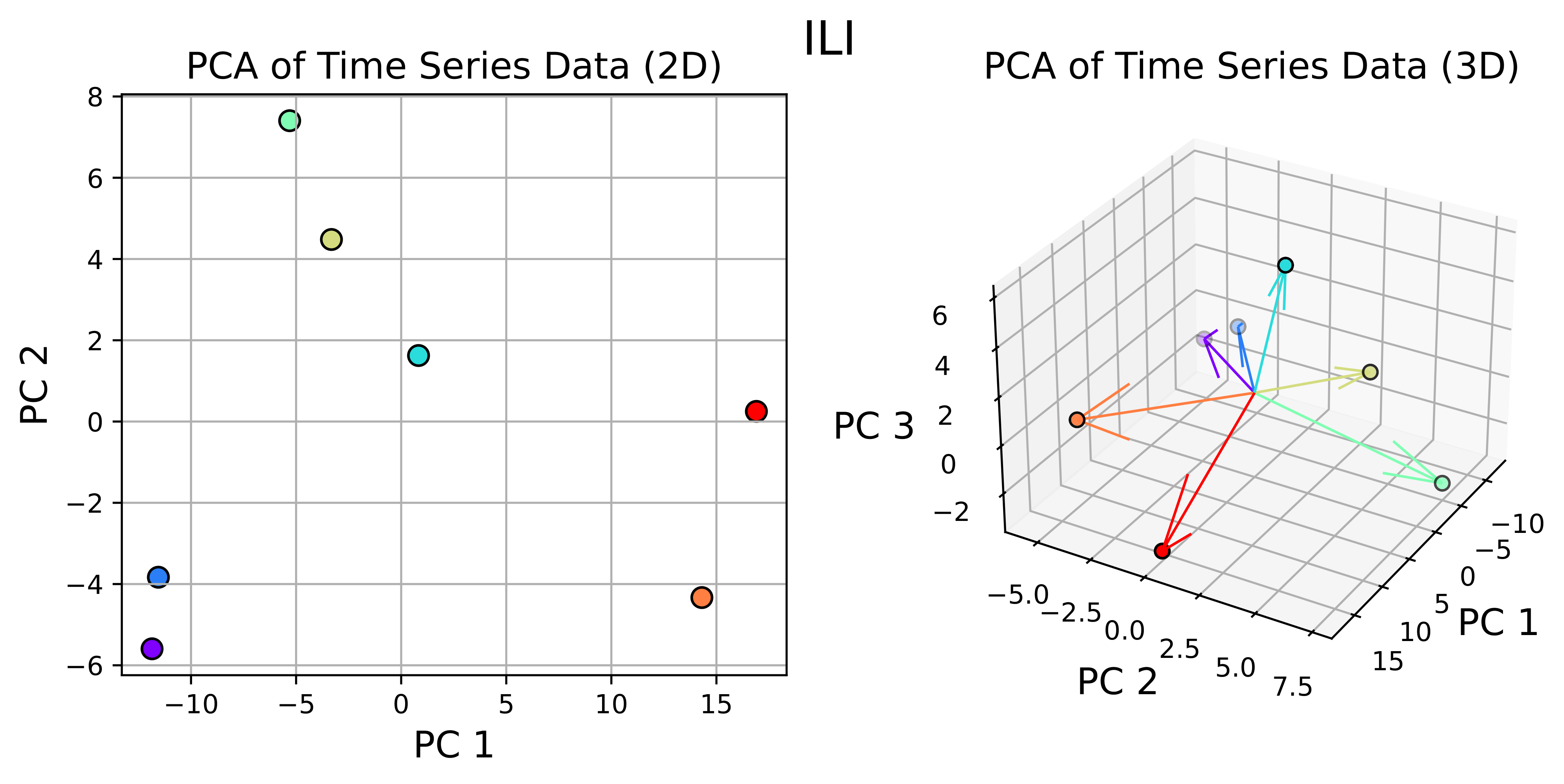} 
    \caption{2D (left) and 3D (right) PCA applied to Exchange, Traffic, and ILI datasets. PCA assists with highlighting variate direction and similarity.}
    \label{fig:PCA_2}
\end{figure}

The figures \ref{fig:PCA_1} and \ref{fig:PCA_2} in the subsection present the results of Principal Component Analysis (PCA) applied to a time series dataset. The left subplot in each row displays a 2D scatter plot of the first two principal components, highlighting the variance captured in two dimensions. Each point represents a data sample, colored to indicate different observations.

The right subplot shows a 3D scatter plot of the first three principal components, providing a more comprehensive view of the data's structure. Vector arrows originating from the origin illustrate the direction and magnitude of each principal component, emphasizing the data's spread in three-dimensional space.

\begin{figure}[!t]
    \centering
    \begin{subfigure}[b]{0.49\textwidth}
        \centering
        \includegraphics[width=\textwidth]{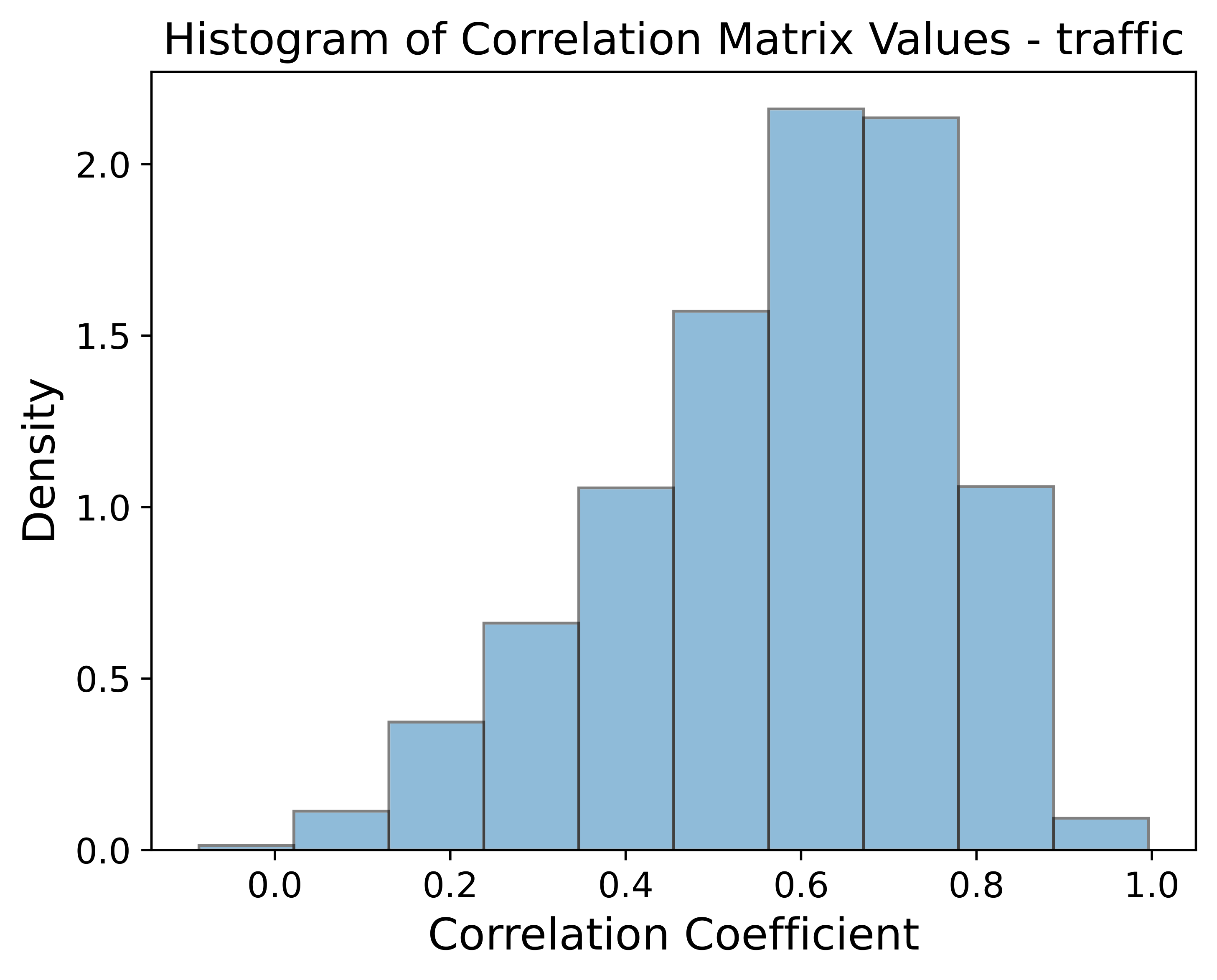}
    \end{subfigure}
    \hfill
    \begin{subfigure}[b]{0.49\textwidth}
        \centering
        \includegraphics[width=\textwidth]{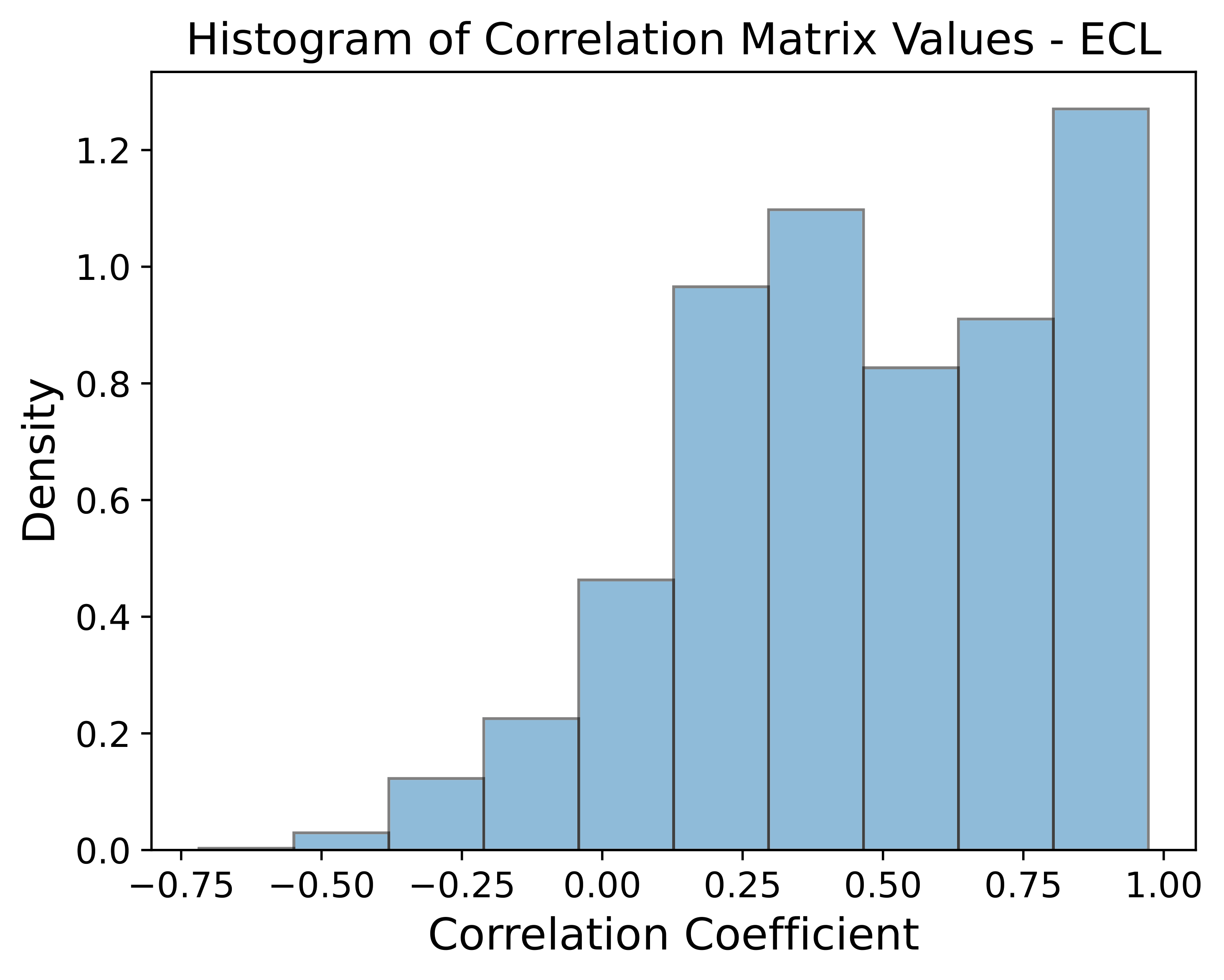}
    \end{subfigure}
    
    \caption{Comparison of the distribution of the correlation coefficient between variates between ECL and Traffic. A skewness closer to 1.0 implies a dataset with many strong linear relationships between variates, therefore, both ECL and Traffic have many variate pairs with a strong linear relationship.}
    \label{fig:corr_histogram}
\end{figure}

\clearpage
\section*{Checklist}


 \begin{enumerate}

 \item For all models and algorithms presented, check if you include:
 \begin{enumerate}
   \item A clear description of the mathematical setting, assumptions, algorithm, and/or model. [\textbf{Yes}. Mentioned in \ref{sec:background}]
   \item An analysis of the properties and complexity (time, space, sample size) of any algorithm. [\textbf{Yes}. Mentioned in \ref{sec:method} and App. \ref{app:mem}.]
   \item (Optional) Anonymized source code, with specification of all dependencies, including external libraries. [\textbf{No}. The code will be provided upon acceptance.]
 \end{enumerate}

 \item For any theoretical claim, check if you include:
 \begin{enumerate}
   \item Statements of the full set of assumptions of all theoretical results. [\textbf{Yes}. See App. \ref{app:grad_prod}]
   \item Complete proofs of all theoretical results. [\textbf{Yes}. See App. \ref{app:grad_prod}]
   \item Clear explanations of any assumptions. [\textbf{Yes}. See App. \ref{app:grad_prod}]     
 \end{enumerate}

 \item For all figures and tables that present empirical results, check if you include:
 \begin{enumerate}
   \item The code, data, and instructions needed to reproduce the main experimental results (either in the supplemental material or as a URL). [\textbf{No}. The code will be provided upon acceptance, but all the experimental details are described in App. \ref{app:eval_setup}.]
   \item All the training details (e.g., data splits, hyperparameters, how they were chosen). [\textbf{Yes}. See in App. \ref{app:eval_setup}.]
         \item A clear definition of the specific measure or statistics and error bars (e.g., with respect to the random seed after running experiments multiple times). [\textbf{Yes}. see in App. \ref{app:extened_results}]
         \item A description of the computing infrastructure used. (e.g., type of GPUs, internal cluster, or cloud provider). [Yes. see in App. \ref{app:eval_setup}.]
 \end{enumerate}

 \item If you are using existing assets (e.g., code, data, models) or curating/releasing new assets, check if you include:
 \begin{enumerate}
   \item Citations of the creator If your work uses existing assets. [\textbf{Yes}. Citation are present for relevant models and datasets.]
   \item The license information of the assets, if applicable. [\textbf{Not Applicable}]
   \item New assets either in the supplemental material or as a URL, if applicable. [Yes. See App. \ref{app:eval_setup}.]
   \item Information about consent from data providers/curators. [\textbf{Not Applicable}]
   \item Discussion of sensible content if applicable, e.g., personally identifiable information or offensive content. [\textbf{Not Applicable}]
 \end{enumerate}

 \item If you used crowdsourcing or conducted research with human subjects, check if you include:
 \begin{enumerate}
   \item The full text of instructions given to participants and screenshots. [\textbf{Not Applicable}]
   \item Descriptions of potential participant risks, with links to Institutional Review Board (IRB) approvals if applicable. [\textbf{Not Applicable}]
   \item The estimated hourly wage paid to participants and the total amount spent on participant compensation. [\textbf{Not Applicable}]
 \end{enumerate}

 \end{enumerate}

\end{document}